\documentclass{article} 
\usepackage{iclr2026_conference,times}

\usepackage{amsmath,amsfonts,bm}

\def\eqref#1{equation~\ref{#1}}

\def\1{\bm{1}}

\DeclareMathAlphabet{\mathsfit}{\encodingdefault}{\sfdefault}{m}{sl}
\SetMathAlphabet{\mathsfit}{bold}{\encodingdefault}{\sfdefault}{bx}{n}

\usepackage{hyperref}
\usepackage{url}
\usepackage{booktabs}
\usepackage{array}
\usepackage{multirow}
\usepackage{graphicx}
\usepackage{amsmath}
\usepackage{amssymb}
\usepackage{xcolor}
\usepackage{colortbl}
\definecolor{deltagain}{HTML}{80ED99}  
\definecolor{deltaloss}{HTML}{FF7171}  
\definecolor{trainframe}{HTML}{B5443F}  
\usepackage{listings}
\usepackage{enumitem}
\usepackage{tabularx}
\usepackage{placeins}  

\lstdefinestyle{toolschema}{
  basicstyle=\ttfamily\footnotesize,
  breaklines=true,
  breakatwhitespace=false,
  columns=fullflexible,
  keepspaces=true,
  showstringspaces=false,
  frame=single,
  rulecolor=\color{black!20},
  framesep=3pt,
  xleftmargin=0pt,
  aboveskip=4pt,
  belowskip=6pt,
  escapeinside={(*@}{@*)}
}
\newcommand{\diff}[1]{\textcolor{blue}{#1}}

\newcommand{\taub}{$\tau$-bench}
\newcommand{\tautwo}{$\tau^2$-bench}

\title{Action-Space Shaping for LLM Agents: \\
Measuring and Mitigating Tool-Schema Bias}

\author{Anonymous Authors}

\author{%
Yinhong Liu$^{1,2}$\thanks{Project repo is available at: \url{https://github.com/williamLyh/ToolSchemaSpace}} \quad Zhili Tan$^{3}$ \quad Zilin Wang$^{4,2}$ \quad Zhijiang Guo$^{4,5}$\\
\textsuperscript{1}University of Cambridge\quad
\textsuperscript{2}Yinwang \quad
\textsuperscript{3}Huawei \quad
\textsuperscript{4}LARK, HKUST (GZ) \quad
\textsuperscript{5}HKUST \\
\small{\texttt{yl535@cam.ac.uk}} \quad
\small{\texttt{chiliktam@gmail.com}} \\
\small{\texttt{zwang374@connect.hkust-gz.edu.cn}} \quad
\small{\texttt{zhijiangguo@hkust-gz.edu.cn}}
\vspace{-2mm}
}

\iclrfinalcopy 
\renewcommand{\headrulewidth}{0pt} 

\begin{document}

\maketitle
\begin{abstract}
Large Language Models (LLMs) have shown strong performance on tool-use agentic tasks when given a fixed tool schema. Yet a tool schema is not the action space of an agent; it is merely one interface representation of it. The same executable action can be exposed through many different, functionally equivalent tool definitions, and an agent that has truly learned a task should behave consistently across them. We show that current agents often do not, a phenomenon we term \emph{schema bias}.
To study this systematically, we introduce an executable transformation framework that rewrites a native tool schema using nine operators, including merging and splitting tools, altering how a single tool is expressed, and distributing one action across several dependent calls. The tasks, executable actions, and reachable states remain fixed, so any change in success is attributable to the interface alone. Evaluating eleven LLMs, including two closed models, on up to 32 schema variants, we ask how large schema bias is, how it manifests, whether the difficulty of a schema variant can be predicted without a full evaluation, and whether training removes it.
We find that schema bias is substantial even for the newest models: success rates range from complete failure to $97\%$ depending solely on the schema. 
To reliably estimate schema difficulty, it requires running a small sample of the target queries. Training repairs a schema variant only when that variant appears in the training data.




\end{abstract}

\section{Introduction}

\begin{figure}[t]
\begin{center}
\vspace{-7mm}
\includegraphics[width=\linewidth]{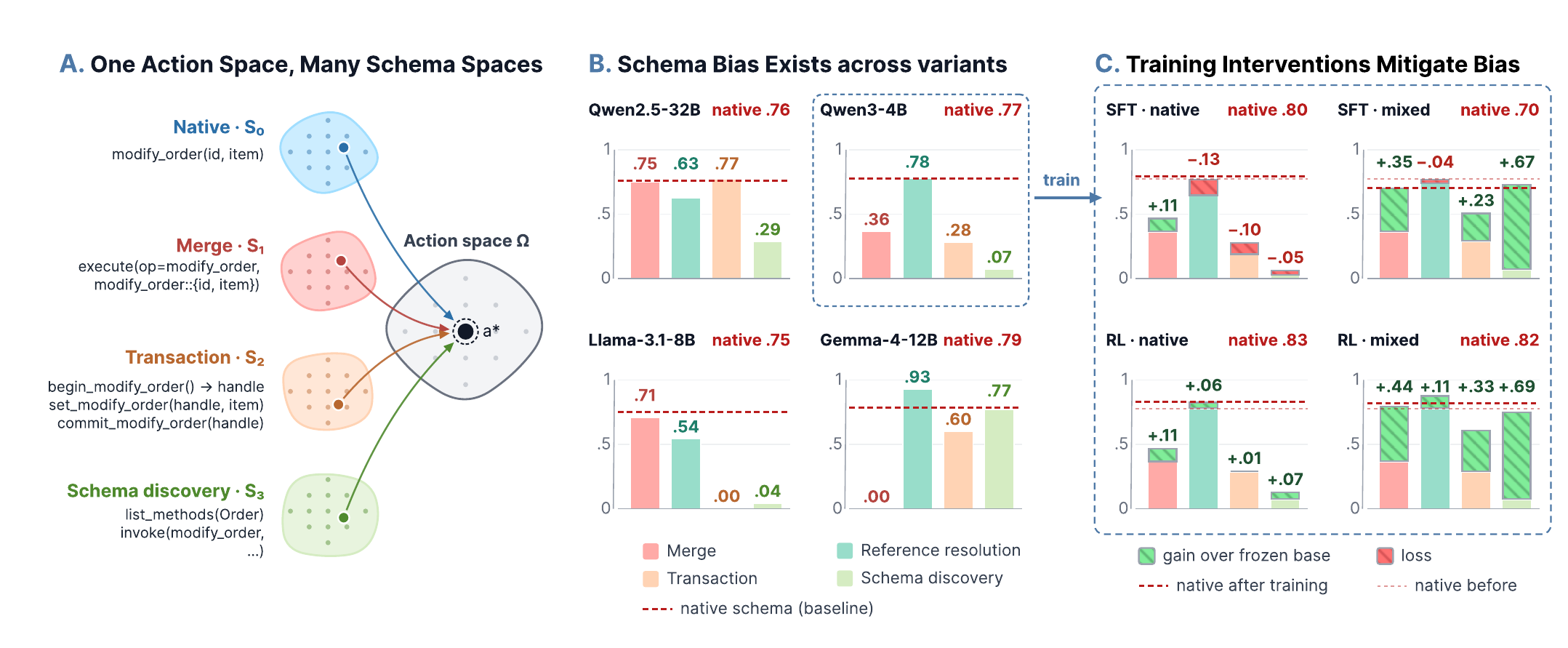}
\end{center}
\vspace{-5mm}
\caption{\textbf{Schema variants, schema bias, and training mitigation.}
\textbf{(A)} All schema variants $S_i$ are just different
representations of the same action space $\Omega$: a call in any of them
decodes to the same native action $a^*$ and reaches the same final state.
Each space is shown with an abbreviated schema for one action.
\textbf{(B)} Schema bias is large and model-specific. Success rates of four
LLMs under four representative variants; the dashed line is each model's
native-schema score. The same variant can match native for one model and
collapse to zero for another (full grid in Figure~\ref{fig:landscape}).
\textbf{(C)} Qwen3-4B after SFT or RL on native-only or
variant-mixture data. Hatched segments are
gains (green) or losses (red). Native-only data leaves most of the bias under
either method; RL on variant-mixture data mitigates bias while
keeping the native score (\S\ref{sec:rq4}).}
\label{fig:teaser}
\vspace{-2mm}
\end{figure}

Large language model (LLM) agents interact with external environments through
tool calls. The interface presented to an agent is typically specified by a
\emph{tool schema}: a declaration of callable functions, their names and
descriptions, argument structures and types, class organization. Although this schema is often treated as part of the environment, it is more accurately as one
representation of the environment's action space.
For example, as shown in Figure 1, one native tool \emph{modify\_order} can be expressed via many different but equivalent approaches.
These alternatives differ in names, factorization, and interaction length, yet they admit a mechanical
translation to the same native actions and reach the same final states. They
are therefore different representations of one action space, not different
tasks. An agent that has learned the task should behave consistently across
them. Figure~\ref{fig:teaser} also illustrates four such representations, the
model-dependent gaps they produce, and how training moves them.

This non-uniqueness matters for research and practice alike. Benchmarks
evaluate agents under a single schema chosen by their designers, implicitly
treating one representation as canonical; reinforcement-learning environments
commit to one factorization of the action space; and deployed systems merge,
split, or reorganize their tools mostly for engineering reasons. If agent
behavior depends on these choices, benchmark scores and training gains partly
reflect an arbitrary interface rather than task difficulty or transferable
capability.

This points to a distinct dimension of agent generalization. Beyond
robustness to prompt wording, observation noise, or distribution shift, a
capable agent should be robust to equivalent representations of the same action
space. We call systematic performance variation across such representations
\emph{schema bias}, and argue that robustness to it should be treated as a
property of the agent rather than left to careful benchmark design.
We study schema bias through four research questions: (RQ1) how large and
systematic is schema bias; (RQ2) how does the representation change the way
agents fail; (RQ3) can the difficulty of a schema be predicted without a full
evaluation; and (RQ4) can training reduce schema bias, and does the reduction
transfer?

We observe that current agents exhibit large and structured schema bias.
Success ranges from complete failure to near-perfect depending only on the
schema, the most difficult representation differs across model families, and
neither the newest open models nor two closed models are immune. Failures follow
the variant: a given schema tends to break different models in the same way,
while how much it costs depends on the model. Reliably ranking schema variants
by difficulty requires running a small sample of the target queries. Training
repairs a variant only when that variant appears in the training data; on-policy
RL does so without the tax that SFT imposes on other variants, and the gains
extend to new combinations of trained schema changes but not to new kinds of
change.
Our contributions follow the progression from measurement to mitigation.
\begin{itemize}[leftmargin=*, itemsep=1pt, topsep=2pt, parsep=1pt]
    \item \textbf{Framework:} we introduce an executable framework of nine operators that transform a native tool schema, from merging and splitting tools to multi-call protocols, and verify every variant against native actions and final environment states. 
    \item \textbf{Measurement:} we measure schema bias in eleven LLMs across up to 32 schema variants.
    \item \textbf{Diagnosis:} we characterize how agents fail under each variant with an automatic failure taxonomy, and evaluate ways to estimate the difficulty of a schema variant, from probes that need no task queries to small samples of the target queries.
    \item \textbf{Mitigation and transfer:}  we compare training-free methods, supervised fine-tuning, and on-policy RL for reducing schema bias, and test how far their gains transfer to held-out domains, held-out schema changes, and real benchmarks.
\end{itemize}
Together, these results position robustness to the schema representation as a
separate dimension of agent generalization: evaluations should report
performance over a small set of verified-equivalent schemas, and training claims should be tested beyond the interface on which they were trained.


\section{Related Work}
\label{sec:related}

\textbf{Tool-use benchmarks fix one schema per environment.} Benchmarks such as \taub{}/\tautwo{} \citep{yao2024taubench,
barres2025tau2}, AppWorld \citep{trivedi2024appworld}, BFCL \citep{bfcl2024,
patil2023gorilla}, ToolSandbox \citep{lu2024toolsandbox}, and ACEBench
\citep{chen2025acebench} evaluate each environment through one fixed tool
schema (see \citet{mohammadi2025survey} for a survey of agent evaluation). These schemas contain consequential representation choices: \taub{}'s
POMDP action space is one parameterization of its domain, AppWorld groups 457
APIs into application classes, and the Model Context Protocol
\citep{anthropic2024mcp} supports runtime disclosure of tool definitions. Our
work treats such choices as variables and evaluates agents across
verified-equivalent representations of the same executable action space.

\textbf{Previous robustness studies perturb tool surface representations, not how its actions
are factored.} Prior work on interface sensitivity primarily varies how a fixed
tool is described. MetaTool \citep{huang2024metatool} shows that descriptions may need
to be adapted to the downstream LLM, while robustness benchmarks such as
RoTBench \citep{ye2024rotbench} corrupt tool definitions. Closest to our
setting, \citet{lee2026patool} observe that small models hallucinate tool names
that follow their pretraining conventions rather than the given schema, and
rename tool components to match those conventions without retraining. Renaming
is one of our operators; we measure it alongside eight others and ask which
models and training methods remove the resulting bias. We study a different
source of variation: invertible reparameterizations that preserve the
available actions and reachable environment states.

\textbf{How actions are represented is known to matter in reinforcement
learning, but has not been tested for tool-using agents.} Our framing is related
to action-space design in reinforcement learning, where
action removal, discretization, factorization, and temporal abstraction affect
learning efficiency \citep{kanervisto2020action,vinyals2017starcraft,
sutton1999between}. We extend this perspective to tool-using language agents
and ask whether a trained model behaves consistently across equivalent action
representations. This differs from prompt-format sensitivity
\citep{sclar2023quantifying,mizrahi2024state}, which changes surrounding text
without changing the factorization of the action space; matched surface
controls help us separate these effects.

\textbf{Function-calling training scales data and methods around a fixed
schema.} Function-calling data and training provide candidate interventions for
schema bias. Recent pipelines synthesize executable tool environments or simulated app
interactions to generate tool-use training data at scale
\citep{xu2026envfactory,liu2024toad}, and \citet{tam2026smith} train a single
policy to both write and use tools, so that the schemas a model creates are ones
it can call. More broadly, self-improving training loops now optimize each stage
of training, down to the design of the training environment itself
\citep{chen2026trainee}; yet the tool schema is treated as a fixed part of the
environment, and whether a trained model has overfit to one specific interface
is rarely examined. APIGen/xLAM
\citep{liu2024apigen}, ToolACE \citep{liu2024toolace}, and Granite
function-calling \citep{abdelaziz2024granite} synthesize or granularize training
data, but do not evaluate invariance across equivalent schemas. We compare
supervised fine-tuning with on-policy GRPO \citep{shao2024deepseekmath} and
measure both in-distribution repair and transfer to held-out domains. The
schema axes also reflect long-standing API-design choices, including
fine-grained REST versus coarse RPC interfaces and opaque identifiers versus
natural keys \citep{fielding2000rest}; our experiments quantify their effects
on agent performance.

\section{An Executable Schema Conversion Framework}
\label{sec:instrument}

\subsection{Framework components}

The conversion framework has four components. First, a \emph{native environment} provides
an executable schema $S_0$ and an execution interface that applies actions to
the environment state. Each native invocation is normalized into an operation
and its typed arguments; we call this record a \emph{native action}. Second, a
library of \emph{transformation operators} rewrites $S_0$ into a variant schema $S_i$.
Third, an \emph{execution adapter} decodes schema-specific tool calls into native actions. 
Finally, a \emph{variant validator} checks that $S_i$ executes the same native
actions and reaches the same environment states as $S_0$. Thus, a
transformation changes how an agent expresses a tool call, but not the actual executable decisions.

\subsection{Schemas, native trajectories, and interaction traces}

We distinguish three objects. A \emph{native action trajectory}
$a^*=(a_1,\ldots,a_K)\in\Omega^*$ records the semantic actions required by a
task, which do not depend on the schema variant.
A \emph{schema} $S_i$ specifies how those actions can be expressed. 
An \emph{interaction trace} $\tau_i$ records how an agent actually invokes tools under the given schema variant $S_i$. 
Let $\mathcal{T}_i$ be the set of completed traces under $S_i$ that can achieve the same task. The execution decoder
maps an observed trace to the native actions that it executes:
\[
D_i:\mathcal{T}_i\rightarrow\Omega^*,\qquad
\tau_i\mapsto\hat a_i.
\]
The decoder can be many-to-one, which means different agent
traces could execute the same native trajectory. 
For example, a model may set transaction arguments in different orders or recover from a rejected call before committing the same action.



\subsection{Transformation operators}
\label{sec:axes}
A transformation operator $\mathcal{O}$ is a class of mechanism that rewrites
the native schema; a concrete rewrite fixes a \emph{method}
$m$, the parameters that instantiate that mechanism. A schema variant is
\begin{equation}
S_i=\mathcal{O}(m, S_0),
\label{eq:variant}
\end{equation}
For example, the operator merge could have different aggregation strategies, e.g.\ by semantic class, by domain, or at random, defined by the \emph{method} $m$. The main figures report only selected representative variants.
We define nine operators, grouped into three categories by what they change (Table~\ref{tab:axes}). 
Operators in the first two categories change only how the tools are presented,
so each native action is still a single tool call; operators in the third
category change the structure of the interaction, so a single native action is
expanded into several dependent tool calls.
Full variant evaluations are shown in Appendix~\ref{sec:app-variants}.


\begin{table*}[t]
\vspace{-10pt}
    \caption{The schema operator space, grouped by what each mechanism changes.
    Each operator is a class of mechanism; a schema variant fixes its method
    (Eq.~\ref{eq:variant}).}
    \label{tab:axes}
    \vspace{6pt}   

    \centering
    \scriptsize
    \setlength{\tabcolsep}{5pt}
    \renewcommand{\arraystretch}{1.15}
    \begin{tabularx}{\textwidth}{@{}
      >{\raggedright\arraybackslash}p{1.7cm}
      >{\raggedright\arraybackslash}p{2.6cm}
      >{\raggedright\arraybackslash}X@{}}
    \toprule
    \textbf{Family} & \textbf{Operators} & \textbf{Mechanisms} \\
    \midrule
    \multirow[t]{2}{*}{Single-call}
    & \emph{merge} \newline
      \emph{split}  \newline
    & \textbf{Tool-set partitioning}.
      Re-designs tool boundaries over the native operations: several
      functions are fused into one tool behind a discriminator, or one
      tool is split by a condition on its arguments.
      Representative methods (Figure~\ref{fig:landscape}): \emph{fully merged}
      and \emph{class dispatch} for merge; \emph{fully split} and
      \emph{interval split} for split. \\
    \cmidrule(l){2-3}
    & \emph{nest arguments} \newline
      \emph{rename} \newline
      \emph{strip descriptions} \newline
      \emph{reorder arguments}
    & \textbf{Per-tool representation}.
      Rewrites the surface form of a single tool expression: parameter nesting structure, tool and
      parameter identifiers, presence of descriptions, and parameter order. 
      Representative methods: \emph{nested args} for nest and \emph{namespaced names} for
      rename; strip descriptions and reorder arguments have a single method each. \\
    \midrule
    Multi-call
    & \emph{transaction} \newline
    \emph{reference resolution} \newline
    \emph{schema discovery}
    & \textbf{Cross-call protocol}.
      Spreads one native action over several dependent calls whose
      state must be carried across turns: 
      \emph{transaction} replaces one native action with multiple intermediate calls.
      \emph{reference resolution} requires an earlier additional reference resolving call. 
      \emph{schema discovery} requires schema query before tool invoke.\\
    \bottomrule
    \end{tabularx}
\end{table*}


\paragraph{Tool-set partitioning} changes the grouping or splitting of the native functions.
The aggregation operator \emph{merge} moves a tool-call decision from the function name into an argument. Which functions go together is the method $m$: \emph{fully merged}
combines all functions of a domain into one tool, \emph{class dispatch} groups the functions of
each class across the whole catalog, and further methods group functions by semantic similarity
or at random, merge only part of a domain, or change how the merged tool takes its arguments.
The splitting operator \emph{split} breaks a tool apart by the value range of one argument:
\emph{fully split} bakes every enumerable argument into the tool name, and \emph{interval split}
does the same for a non-enumerable argument by cutting its range into intervals.
Figure~\ref{fig:landscape} includes only these two merge methods and two split methods; the
remaining methods are described in Appendix~\ref{sec:app-methods} and evaluated in
Appendix~\ref{sec:app-variants}.

\paragraph{Per-tool representation} operators modify a tool in place. We define four:
\emph{nest} wraps a tool's arguments into one nested object;
\emph{rename} changes the tool identifiers (function names), either by prefixing them with
their class name (namespaced names) or by replacing them with opaque identifiers;
\emph{strip descriptions} and \emph{reorder arguments} change only the textual presentation.
Nest changes the parameter-schema tree, whereas the other three change its labels or
ordering; none of them changes the selected native operation or the minimum call path.

\paragraph{Cross-call dependency} operators expand the same native action into a sequence
of tool calls with sequential dependencies: a later call consumes what an earlier call
returned. We define three. \emph{Transaction} replaces one native action with several
intermediate calls that open a transaction, write the arguments one call at a time, and
commit; such protocols are usually motivated by external requirements such as formality,
validation, or auditability. \emph{Reference resolution} removes one argument's value from
the tool and adds a resolver: the agent first resolves the user-provided value to a handle
and then passes that handle in place of the value. \emph{Schema discovery} requires the
agent to first list a class's methods and argument names, and to call only with the names
it was given. 

\subsection{Validating schema variants}

We compare each schema variant $S_i$ with the native schema $S_0$ on the same task. 
Each task specifies a user query and a target native trajectory $a^*$.
The operator maps $a^*$ to a variant trace $\tau_i^*$. During variant validation, the execution adapter runs trace $\tau_i^*$ and reconstructs native actions. 
$S_i$ passes validation only if the reconstructed actions match $a^*$.
The schema transformation framework is not tied to a particular benchmark. We also
apply the same framework to both our synthetic controlled environment and
open-sourced tool-use benchmarks: \tautwo{}, BFCL, AutomationBench, and MCP-Atlas.
\section{Experimental Setup}
\label{sec:setup}
\begin{figure*}[t]
\vspace{-5mm}
\centering
\includegraphics[width=\textwidth]{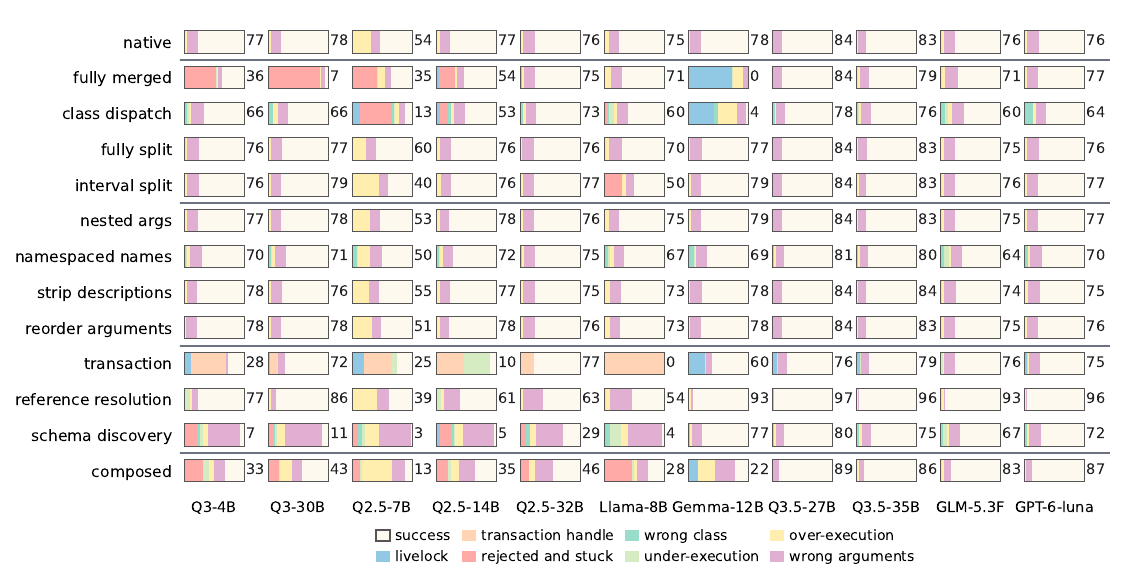}
\caption{\textbf{The model-by-schema performance grid: exact success and failure mode composition.} 
Each cell is one model (column) on one verified-equivalent schema (row), 2{,}748 identical tasks. 
The number is the exact success rate (\%), represented as the white portion of composition bar.
The coloured segments are the absolute rates of the automatically detected
failure modes analysed in \S\ref{sec:rq2}. 
Row labels are the representative variant names following Table~\ref{tab:axes}.
The two rightmost columns are closed models evaluated through their APIs.
The remaining variants are reported in Appendix Figure~\ref{fig:app-variants}.}
\label{fig:landscape}
\end{figure*}

The experiments follow the four research questions: we first measure a
model-by-schema performance landscape (RQ1) and analyze its failure modes
(RQ2), then test ways to estimate schema difficulty cheaply (RQ3), and finally
train models to reduce schema bias and test whether the gains transfer (RQ4).
Every comparison within an environment holds the tasks, the native execution
interface, and the scorer fixed.

In brief, the controlled synthetic environment has 12 domains, 168 native
operations, and 2{,}748 task queries; \tautwo{}, BFCL,
AutomationBench, and MCP-Atlas serve as real-benchmark validation. Nine open-weight models from the Qwen2.5, Qwen3, Qwen3.5, Llama-3.1,
and Gemma-4 families \citep{qwen2024qwen25,qwen2025qwen3,qwen2026qwen35,dubey2024llama,gemma2026gemma4}
are evaluated on all 32 variants and two closed models, gpt-6-luna
\citep{openai2026gpt6} and GLM-5.3-Flash \citep{glm2026glm5}, on the thirteen
representative variants of
Figure~\ref{fig:landscape}. An episode succeeds when the native actions it
executes match the gold actions exactly, and calls that violate the active
schema are rejected and returned to the model as an error it can recover from.
Appendix~\ref{sec:app-env} gives the full setup: the synthetic domains, query
statistics, and query construction, the real-benchmark adapters, the models and
their settings, and the evaluation protocol.

\section{RQ1: Measuring Schema Bias}
\label{sec:rq1}

Our first research question asks how much task success varies across
equivalent schema variants, and whether that variation follows reproducible
patterns rather than evaluation noise. We derive 32 verified-equivalent
variants (the native schema and 31 transformed variants) from one native
action space and evaluate all of them on nine open-weight models, and the
representative variants also on two closed models.
Figure~\ref{fig:landscape} reports the representative variants (the rest are
in Appendix Figure~\ref{fig:app-variants}) together with how the failed
episodes are distributed over failure modes, which \S\ref{sec:rq2}
analyses. With tasks and native semantics fixed, success still ranges from
complete failure to near-perfect depending only on the schema, and the same
variant can be catastrophic for one model and harmless for another. No single
variant wins on every model, and each model family is sensitive to different
representations; the newest Qwen3.5 generation is the most robust but not
immune. The closed models are not immune either: equivalent schemas move both
by 32 points, and their weak spots are the variants that expose a large
catalog (class dispatch and namespaced names) rather than the cross-call
protocols that break most open models. The rest of this section organises this variation into recurring
patterns by operator family.


\begin{figure}[t]
\vspace{-5mm}
\centering
\includegraphics[width=0.925\linewidth]{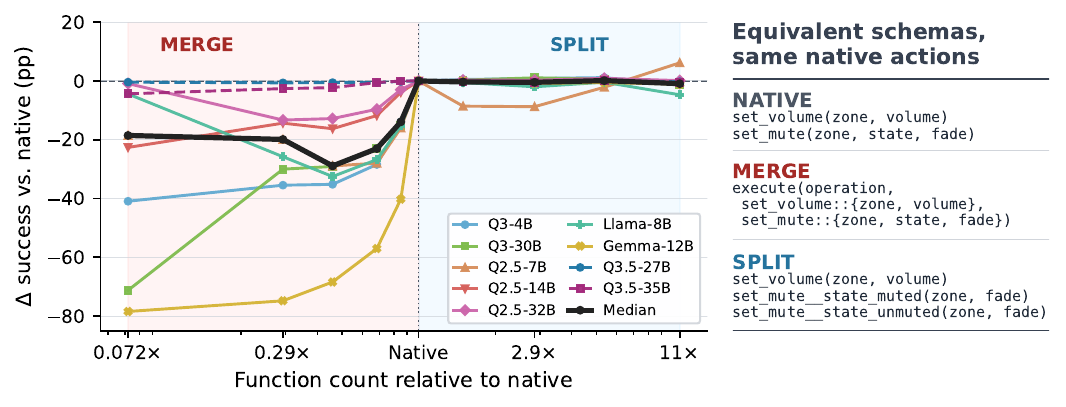}
\vspace{-5mm}
\caption{\textbf{Merging is generally more harmful than splitting.}
Left: colored lines show each model's change from its native-schema success
rate. The black line is the median across the nine models. 
The x-axis represents ratio of function count between variant and native schema. We sample ten variants to span the granularity ladder from full merge to full split.
Each ratio is the query-weighted mean across 12 schema domains. 
Right: a compact and abbreviated schema example shows how merging and enum splitting transform the native schema.}
\label{fig:rq1_granularity}
\end{figure}
\subsection{Recurring structural patterns}

Results across the nine-model by 32-variant grid are model-dependent, but
recurring patterns organize much of the variation. We group them by the
operator family of Table~\ref{tab:axes} and, because no pattern holds for
every model, report the main trend together with its exceptions. All results
use the default evaluation protocol (Appendix~\ref{sec:app-protocol}).

\textbf{Tool-set partitioning:} \emph{merging is generally more harmful than
splitting.} Tool Merging and splitting are two opposite and complementary directions, but they do not show similar challenges to LLMs.
Figure~\ref{fig:rq1_granularity} compares a representative ten-step
granularity ladder across all nine models, spanning $0.07\times$ to
$10.6\times$ the native function count. Splitting largely preserves
native-schema performance, whereas fully merging tools reduces success by a median of 19 points.
\emph{Different merge methods cause further challenges.} The merged tools
above use the flat argument structure; Appendix~\ref{sec:app-methods} re-expresses
the same merged arguments in the nested and union forms. No structure is safe
for every model: the nested form costs every model 19 to 73 points, and even the
otherwise robust Qwen3.5 models, which lose under 7 points on the flat merge,
fall to $0.20$--$0.25$ on it.

Besides, which functions are merged together matters
as well: at a fixed granularity, grouping semantically related operations is
consistently better than random or deliberately mismatched grouping, and the
gap widens as the groups get larger. Mixing argument structures within one
schema variant adds a further cost: for all but one of the nine models, the
mixed catalog scores below the average of its constituent structures
(Appendix~\ref{sec:app-grouping}, \ref{sec:app-mixtures}).

\textbf{Per-tool surface changes pose limited challenges for most models.}
Nesting the arguments, removing descriptions, permuting argument order, and
namespacing tool names leave every model close to its native score; the one
exception is that opaque function names hurt the smallest Qwen2.5 model, and
that sensitivity vanishes with scale (Appendix~\ref{sec:app-surface}).

\textbf{Cross-call protocols are the hardest operators for most models.}
Spreading one action over several dependent calls produces the largest
losses in the landscape: the transaction protocol and schema discovery
collapse most earlier-generation models, often to near zero, and only the newest
generation absorbs them. Reference resolution is the exception in the other
direction: the models that handle the extra resolver call score at or above
their native rate, so a cross-call protocol is not harmful in itself, but
whether a model follows it is model-specific (\S\ref{sec:rq2} analyses the
failure modes).

\subsection{Real tool-use benchmark validation}
\label{sec:rq1_real}

To test whether schema bias survives outside the synthetic grid, we rewrite the
tools of four real benchmarks (\tautwo{} airline and retail, BFCL,
AutomationBench, and MCP-Atlas) with eleven of the representative variants, and
evaluate Qwen3.5-27B and Qwen3-4B with no self-correction
(Appendix~\ref{sec:app-real-proxy}, Table~\ref{tab:real_proxy}). Schema bias
persists, and the costly variants are the same as on the synthetic grid. Fully
merged, class dispatch, and composed cost Qwen3.5-27B 0.26 to 0.67 on BFCL,
AutomationBench, and MCP-Atlas, and fully merged costs it 0.30 on \tautwo{}
retail, while the variants that change only a tool's own representation stay
within 0.10 of native. The smaller model loses more: on BFCL, fully merged,
class dispatch, schema discovery, and composed take Qwen3-4B from 0.33 to 0.03
or below. \tautwo{} airline, with 50 tasks, separates the variants least.


\section{RQ2: Failure Modes Under Equivalent Schemas}
\label{sec:rq2}

We analyze how schema representation changes the type of error an agent
makes. Across failed episodes we observe seven recurring failure
modes, listed in Table~\ref{tab:signatures}: four explicit ones (livelock,
transaction handle, rejected and stuck, wrong class) and three silent ones
(under-execution, over-execution, wrong arguments). The labels are
automatic; in a blinded audit (Appendix~\ref{sec:app-audit}), three LLM
proxies reach a majority label on 88\% of sampled failed episodes.

\begin{table*}[t]
\vspace{-5mm}
    \caption{The seven failure modes. Every failed episode is assigned the
    first mode whose rule fires. The first four surface as explicit runtime errors
    or routing errors; the last three are silent failure: the episode ends with the
    wrong set of native actions or the wrong argument values.}
    \label{tab:signatures}
    \vspace{4pt}

    \centering
    \scriptsize
    \setlength{\tabcolsep}{4pt}
    \renewcommand{\arraystretch}{1.05}
    \begin{tabularx}{\textwidth}{@{}
      >{\raggedright\arraybackslash}p{2.4cm}
      >{\raggedright\arraybackslash}X
      >{\raggedright\arraybackslash}X@{}}
    \toprule
    \textbf{Failure mode} & \textbf{Rule} & \textbf{What went wrong} \\
    \midrule
    Livelock & The episode hits the turn limit. & The agent loops on retries or clarifications without committing an action. \\
    Transaction handle & A transaction flag fires: a write or commit names a transaction that was never opened, an opened transaction is never committed, a commit precedes the required writes, or a transaction is committed twice. & The begin / write / commit protocol is broken. \\
    Rejected and stuck & At least one call was rejected as off-schema and fewer native actions executed than required. & The agent calls a name or argument the schema does not expose and cannot recover. \\
    Wrong class & A call is routed through a merged tool or namespace of the wrong class. & The operation is chosen from the wrong group. \\
    \midrule
    Under-execution & Fewer native actions executed than the task requires, with no rejection. & Required steps are skipped. \\
    Over-execution & More native actions executed than the task requires. & Extra or repeated actions are taken. \\
    Wrong arguments & The required actions execute with at least one wrong argument value. & The right functions are called with the wrong values. \\
    \bottomrule
    \end{tabularx}
\vspace{-3mm}
\end{table*}

\textbf{Failure modes cluster on certain variants and models.}
Figure~\ref{fig:landscape} shows how the failure modes are distributed
across models and representative schemas. The same native tasks produce
different error types as the schema changes. 
Some failure modes are largely consistent across models: fully merged catalogs fail as rejected-and-stuck, the transaction protocol as broken transaction handles, and schema discovery as silent wrong arguments.
Two models stand out: Qwen2.5-7B tends to over-execute under almost every
variant, and Gemma-4-12B livelocks on any merged tool.

\textbf{Some failures invoke correct actions but use the wrong tool calls.}
A model sometimes invokes the correct native action but fails to write it in
the presented schema: the call is off-schema, yet it decodes to the right
native action. We call these \emph{schema-form failures} and measure them by
re-running each variant with off-schema calls executed as the native action
they decode to rather than rejected (Appendix~\ref{sec:app-schema-form}); the resulting delta in
success is the schema-form gap, shown in Figure~\ref{fig:rq2_schema_form}. 

Schema-form failures concentrate on variants that rename or regroup tools and
on the earlier model generations. Fully merged has by far the largest gap,
driven by the Qwen3 and smaller Qwen2.5 models, which keep calling native tool
names; composed variants show a clear gap for every
earlier model except Gemma-4-12B, and on interval split almost all of
Llama-3.1-8B's loss is schema-form. The recovered episodes are mostly
rejected-and-stuck failure modes. The two Qwen3.5 models and
Gemma-4-12B show almost no schema-form failures on any variant.


\section{RQ3: Predicting Schema Difficulty Without a Full Evaluation}
\label{sec:rq3}

An environment designer who exposes a tool schema also decides how hard the
environment is for every agent that will use it. RQ1 showed that this
difficulty is large and model-specific, so before committing to a schema the
designer needs to know how a given model will fare on it. Running the full
query set on every candidate schema answers this reliably but expensively. We
therefore ask what cheaper evidence suffices to rank candidate schemas for a
model, comparing estimators by what they may observe: the model on the schema
alone, or the model with a sample of the target task queries. Each estimator scores every variant, and we measure
how well the scores rank one model's variants by the Spearman correlation
$\rho$ with the full-set success rates, computed within each model and then
averaged over the nine models.

\textbf{Probing the model without task queries is not enough.} We try two
query-free probes, both built from a few dozen simple instructions that are
written automatically from the schema's own tool descriptions and argument
values, without any task queries: the likelihood the model assigns to the
correct call sequence for each instruction, and a schema compliance probe,
which runs the model on the instructions and records how often it follows the
schema's calling convention. Both rank variants only moderately well
($\rho=0.52$ for the likelihood and $0.57$ for the compliance probe). Further to this, we note that schema difficulty is task-conditional, for example, models often follow a protocol on isolated
calls yet fail it inside multi-turn tasks (Appendix~\ref{sec:app-rq3}). Difficulty is therefore a joint property of the model, the schema, and the task queries.

\textbf{A small sample of target queries is the reliable estimate.} Running
the model on about one hundred queries per candidate schema ranks the variants
nearly as the full set does ($\rho=0.81$, and $0.92$ with 800 queries), at a
small fraction of the cost. Even 36 queries, as many as the probe
instructions, reach $\rho=0.72$, so the probes fall short because they omit the
tasks, not because they are small. 
The same holds on the four real benchmarks, where 36 of a benchmark's own tasks
rank the variants nearly as well as the full set (Appendix~\ref{sec:app-rq3}). A
designer should therefore sample from the environment's own task distribution.

\section{RQ4: Reducing Schema Bias Through Training}
\label{sec:rq4}

In this section, we ask whether training can remove schema bias and whether the gains carry over to domains and schemas that were not trained on. We compare three levers of increasing cost on Qwen3-4B: 
Two training-free methods, variant-specific instructions and decoding constraints; Supervised Fine-Tuning (SFT) with LoRA \citep{hu2022lora} and on-policy
GRPO \citep{shao2024deepseekmath}. Both training methods use the same
2{,}000 queries from eight training domains, either on a single schema or
rotating over seven variants including native (\emph{mixed}), and every model is evaluated on the
twelve representative variants, including four held-out domains (settings in
Appendix~\ref{sec:app-rq4}). Table~\ref{tab:rq4-methods} compares all methods on the twelve representative variants.

\begin{table}[t]
\caption{Change in success from each mitigation on Qwen3-4B over the twelve representative
variants of Figure~\ref{fig:landscape}. \emph{Base} is the untrained model; every other column show delta success rate from it.
changes of at least $0.03$ (about 2.5 standard errors) are shaded green or red, darker for larger changes. \emph{Instruction} appends variant-specific calling convention describing instructions to input prompt (Table~\ref{tab:rq4-icl}); \emph{Decoding} forces the first call to be schema-valid via constrained decoding. SFT and RL columns are results averaged from three seeds;
\emph{Mixed} data rotate over seven schema variants. Red frames mark
the variants each training set contains.}
\label{tab:rq4-methods}
\begin{center}
\scriptsize
\setlength{\tabcolsep}{4pt}
\begin{tabular}{@{}l c cc cc cc@{}}
\toprule
& & \multicolumn{2}{c}{Training-free} & \multicolumn{2}{c}{SFT} & \multicolumn{2}{c}{RL (GRPO)} \\
\cmidrule(lr){3-4} \cmidrule(lr){5-6} \cmidrule(l){7-8}
Variant & Base & Instruction & Decoding & Native & Mixed & Native & Mixed \\
\midrule
Native & 0.77 & -- & 0.00 & \setlength{\fboxsep}{1pt}\setlength{\fboxrule}{0.7pt}\fcolorbox{trainframe}{white}{+0.02} & \cellcolor{deltaloss!28}\setlength{\fboxsep}{1pt}\setlength{\fboxrule}{0.7pt}\fcolorbox{trainframe}{deltaloss!28}{$-$0.08} & \cellcolor{deltagain!24}\setlength{\fboxsep}{1pt}\setlength{\fboxrule}{0.7pt}\fcolorbox{trainframe}{deltagain!24}{\textbf{+0.05}} & \cellcolor{deltagain!22}\setlength{\fboxsep}{1pt}\setlength{\fboxrule}{0.7pt}\fcolorbox{trainframe}{deltagain!22}{+0.04} \\
\midrule
\multicolumn{8}{@{}l}{\emph{Tool-set partitioning}} \\
Fully merged & 0.36 & \cellcolor{deltagain!61}+0.28 & \cellcolor{deltagain!63}+0.29 & \cellcolor{deltagain!33}+0.11 & \cellcolor{deltagain!72}\setlength{\fboxsep}{1pt}\setlength{\fboxrule}{0.7pt}\fcolorbox{trainframe}{deltagain!72}{+0.34} & \cellcolor{deltagain!34}+0.11 & \cellcolor{deltagain!87}\setlength{\fboxsep}{1pt}\setlength{\fboxrule}{0.7pt}\fcolorbox{trainframe}{deltagain!87}{\textbf{+0.43}} \\
Class dispatch & 0.66 & +0.01 & $-$0.02 & \cellcolor{deltaloss!58}$-$0.26 & 0.00 & \cellcolor{deltagain!23}+0.05 & \cellcolor{deltagain!24}\textbf{+0.05} \\
Fully split & 0.76 & 0.00 & 0.00 & \cellcolor{deltagain!23}+0.05 & \cellcolor{deltaloss!25}\setlength{\fboxsep}{1pt}\setlength{\fboxrule}{0.7pt}\fcolorbox{trainframe}{deltaloss!25}{$-$0.06} & \cellcolor{deltagain!25}\textbf{+0.06} & \cellcolor{deltagain!23}\setlength{\fboxsep}{1pt}\setlength{\fboxrule}{0.7pt}\fcolorbox{trainframe}{deltagain!23}{+0.05} \\
Interval split & 0.76 & +0.01 & \cellcolor{deltaloss!22}$-$0.04 & \cellcolor{deltaloss!30}$-$0.09 & \cellcolor{deltaloss!27}$-$0.07 & \cellcolor{deltagain!24}\textbf{+0.05} & \cellcolor{deltagain!23}+0.05 \\
\midrule
\multicolumn{8}{@{}l}{\emph{Per-tool representation}} \\
Nested args & 0.77 & +0.01 & +0.01 & +0.02 & \cellcolor{deltaloss!28}$-$0.08 & \cellcolor{deltagain!24}\textbf{+0.06} & \cellcolor{deltagain!22}+0.04 \\
Namespaced names & 0.70 & 0.00 & 0.00 & \cellcolor{deltagain!20}+0.03 & \setlength{\fboxsep}{1pt}\setlength{\fboxrule}{0.7pt}\fcolorbox{trainframe}{white}{+0.01} & \cellcolor{deltagain!26}\textbf{+0.07} & \cellcolor{deltagain!24}\setlength{\fboxsep}{1pt}\setlength{\fboxrule}{0.7pt}\fcolorbox{trainframe}{deltagain!24}{+0.06} \\
Strip descriptions & 0.78 & 0.00 & 0.00 & 0.00 & \cellcolor{deltaloss!31}$-$0.10 & \cellcolor{deltagain!21}\textbf{+0.04} & +0.03 \\
Reorder arguments & 0.78 & +0.01 & 0.00 & +0.02 & \cellcolor{deltaloss!28}$-$0.08 & \cellcolor{deltagain!24}\textbf{+0.06} & \cellcolor{deltagain!22}+0.04 \\
\midrule
\multicolumn{8}{@{}l}{\emph{Cross-call dependency}} \\
Transaction & 0.28 & \cellcolor{deltagain!26}+0.06 & \cellcolor{deltaloss!33}$-$0.11 & \cellcolor{deltaloss!32}$-$0.10 & \cellcolor{deltagain!53}\setlength{\fboxsep}{1pt}\setlength{\fboxrule}{0.7pt}\fcolorbox{trainframe}{deltagain!53}{+0.23} & +0.02 & \cellcolor{deltagain!71}\setlength{\fboxsep}{1pt}\setlength{\fboxrule}{0.7pt}\fcolorbox{trainframe}{deltagain!71}{\textbf{+0.34}} \\
Reference resolution & 0.77 & +0.03 & \cellcolor{deltagain!26}+0.07 & \cellcolor{deltaloss!37}$-$0.13 & \cellcolor{deltaloss!22}\setlength{\fboxsep}{1pt}\setlength{\fboxrule}{0.7pt}\fcolorbox{trainframe}{deltaloss!22}{$-$0.04} & \cellcolor{deltagain!25}+0.06 & \cellcolor{deltagain!33}\setlength{\fboxsep}{1pt}\setlength{\fboxrule}{0.7pt}\fcolorbox{trainframe}{deltagain!33}{\textbf{+0.11}} \\
Schema discovery & 0.07 & \cellcolor{deltagain!90}+0.45 & +0.02 & \cellcolor{deltaloss!23}$-$0.05 & \cellcolor{deltagain!90}\setlength{\fboxsep}{1pt}\setlength{\fboxrule}{0.7pt}\fcolorbox{trainframe}{deltagain!90}{+0.67} & \cellcolor{deltagain!25}+0.06 & \cellcolor{deltagain!90}\setlength{\fboxsep}{1pt}\setlength{\fboxrule}{0.7pt}\fcolorbox{trainframe}{deltagain!90}{\textbf{+0.68}} \\
\bottomrule
\end{tabular}
\end{center}
\end{table}

\textbf{Training-free methods fix habits that one sentence can describe.} The constraint decoding method forces the first call to use tools provided in the schema.
Among all variants, it turns out to help almost only on the
fully merged schema. 
The instruction method, which describes each variant's calling convention in
one sentence, clearly improves fully merged and schema discovery. 
Neither method has a noticeable effect on the other variants.

\textbf{SFT repairs the trained variants but can tax the others.} SFT on
native-schema demonstrations keeps native performance but does not repair the
hard variants, so the repair comes from seeing a variant during training rather
than from more tool-use practice. SFT on mixed data lifts the trained variants
(red frames in Table~\ref{tab:rq4-methods}), while already-robust variants,
trained or not, lose accuracy to different extents. This tax depends
more on how the demonstrations are phrased than on how many there are
(Appendix~\ref{sec:app-rq4}).


\textbf{On-policy RL repairs the trained variants without the tax.} With
either native-only or mixed data, RL keeps every variant at or above its
untrained score. As with SFT, native-only data leaves the hard variants
unrepaired, whereas mixed data gives the largest gains on the trained variants
of any method, making it the most effective mitigation we test. The repair
needs three conditions: the variant must appear in the training data, the model
must be free to move far from its initialization (no KL penalty), and the
environment must reject invalid calls (Appendix~\ref{sec:app-rq4}).



\textbf{Gains transfer to recombinations of familiar schema changes, not to new
ones.} Both training methods keep part of their gain on the four held-out
domains, and RL keeps more than SFT. 
To test generalization across schemas, we run an additional SFT experiment that trains on seven schema variants and evaluates on changes held out from training (Appendix~\ref{sec:app-rq4}): the gains carry over to new combinations of the trained changes and to closely related untrained variants, such as finer interval splits, but not to kinds of operator the model never saw.
On four real benchmarks whose tools we rewrite with the same operators, RL on the mixture gains only on the trained variants ($+0.03$ on average, against $+0.003$ on the untrained ones; Appendix~\ref{sec:app-transfer}, Table~\ref{tab:rq4_real}). Training therefore reduces
the specific schema biases it targets, but does little to make the model robust to
schema changes in general.

\section{Conclusion}

Tool schemas are representations of an agent's action space, yet current
evaluations usually treat one representation as canonical. We introduced an
executable framework that changes the schema while preserving native actions
and final environment states. 
We show that success varies substantially across equivalent schemas, that
this schema bias persists even for the top-ranked models, and that it is
neither a small formatting effect nor a single universal penalty: it varies
systematically with the model and the schema variant.
The failures follow the variant: a given schema tends to break different
models in the same way, while how much it costs depends on the model.
Reliably ranking schema variants by difficulty requires running a small sample
of the target queries.
Training repairs a variant only when that variant appears in the training
data; on-policy RL does so without the tax that SFT imposes on untrained
variants, and the gains extend to new combinations of trained changes but not
to new kinds of schema change.

Finally, we argue that schema bias should be taken into account wherever tool
calling is involved: when designing the tool schemas of an environment, when
training models, and when evaluating tool-calling ability. Robustness to the
schema representation should become a first-class criterion for tool-using
agents.


\subsubsection*{AI Use Statement}
We used large language models to polish the writing of this paper. All
content, analyses, and conclusions are the authors' own, and the authors
checked every edit. Separately, LLMs are the subject of our experiments, and an
LLM rewrote the templated evaluation queries under a grounding check that keeps
the gold calls fixed (Appendix~\ref{sec:app-env}).

\subsubsection*{Ethics Statement}
This work evaluates and trains language-model agents in a synthetic tool-use
environment and on public benchmarks. It involves no human subjects, no
personal or sensitive data, and no real-world side effects: every tool call is
executed in a simulated environment. Closed models were accessed through their
public APIs under the providers' terms of use. We do not foresee specific
ethical risks beyond those common to research on language-model agents.

\subsubsection*{Reproducibility Statement}
Every variant passes reference-trace decoding and state-equivalence checks before
evaluation. Across approximately 50{,}000 reference-trajectory checks, these
tests identified three transformation-implementation errors before model
evaluation and none afterward. We release the checks, environment adapters,
domain specifications, transformation framework, the registry of all 32 schema
variants, and the model--schema evaluation grids. Each reported configuration
uses the full set of 2{,}748 queries with state-based scoring. One full
configuration costs approximately 5 GPU-minutes with data-parallel evaluation on
eight RTX 5090 GPUs, and all grids are resumable by model--schema
configuration. At temperature $0.7$, three independent seeds preserve the
expected high--medium--low ordering for both Qwen3-4B and Llama-3.1-8B; seed
standard deviations are at most $0.008$, with episode-bootstrap confidence
intervals reported per cell. The phrasing-and-volume SFT study completes all
160 planned evaluations (two phrasings, four data sizes, two seeds, and ten
variants), each on the full query set. The two SFT-then-RL runs each use 19{,}200
rollout episodes and 200 optimizer steps, costing 24.2 and 24.5 GH200 GPU-hours.

The ranking analysis uses the full $9{\times}32$ model--variant matrix, with all
288 cells complete and paired at the episode level. The two additional training
studies of Appendix~\ref{sec:app-extra-training} train and evaluate every
planned run (36 GRPO runs and 48 checkpoints, respectively) without selecting
checkpoints or seeds, and bind training data, checkpoints, and evaluation
outputs to recorded digests so that incomplete runs cannot enter the results.


\bibliography{references/references}
\bibliographystyle{iclr2026_conference}

\appendix
\section{Limitations}
\label{sec:app-limitations}

Our framework covers nine operators and 32 schema
variants, but the space of equivalent schemas is larger: other directions, such
as how tool results and errors are reported or how state is exposed across
calls, may reveal biases that our variants do not. The mitigation results are
bounded in the same way. Training removes the bias only for the schema changes
it covers and their combinations, not for new kinds of change, so every new
interface design still has to be evaluated, and if necessary trained for, on
its own; schema bias cannot be fixed once for all. Finally, most measurements
come from a controlled synthetic environment and a finite set of models, so the
size of the effect on other benchmarks and model families remains to be
established.

\section{Experimental setup}
\label{sec:app-env}

This appendix gives the full setup summarized in \S\ref{sec:setup}.

\subsection{Synthetic environment and queries}
\label{sec:app-data}

The controlled environment has twelve synthetic domains with 168 native
operations (Table~\ref{tab:synthetic_domains}). Each domain defines a native
tool schema, namely a system context, callable operations with typed arguments,
and multi-step routines used for compositional queries, from which the tool
catalog and the query sets are derived. The environment implements every
operator of Table~\ref{tab:axes}. The released artifact includes the native
schemas, the query files, and the transformation registry.

The 2{,}748 task queries come in three tiers: 585 \emph{single} queries
requiring one native call, 1{,}747 \emph{compound} queries combining several
independent operations in one utterance, and 416 \emph{compositional} queries
chaining dependent steps within a domain, where later actions reuse entities
established earlier. Table~\ref{tab:synthetic_stats} summarizes the catalog and
the queries. Every tool argument is required and 66\% of arguments are
enumerated, which is what makes the split operators (baking enum values into
tool names) and the merge operators (routing by an \texttt{operation} enum) well
defined. The model--schema landscape uses all 2{,}748 queries in all twelve
domains. For training in \S\ref{sec:rq4} only, eight domains supply the training
queries and four are held out to test transfer (1{,}789 and 959 evaluation
queries). The same query and target native actions are held fixed across all
schema variants.

\textbf{How the queries are built.} The domains are written by hand as a
declarative specification modelled on real device and web interfaces
(home automation, media players, calendars, in-car controls). Gold calls are
sampled programmatically with a fixed seed: a single query calls one tool with
sampled argument values, a compound query samples two to six independent calls
within a domain, and a compositional query instantiates a hand-written
multi-step routine whose later calls depend on earlier ones. Each gold call list
is first rendered as a templated request and then rewritten by
gemini-3.5-flash \citep{google2026gemini35flash} in two passes, both conditioned on the gold calls and never
changing them. The first pass writes a natural request for exactly those calls;
the second adds a persona and situational context, one or two distractor
remarks that map to no tool, filler, and non-linear ordering, which roughly
triples the length. Every concrete gold value must remain recoverable from the
final text: the 49 queries (1.8\%) whose grounding falls below $0.9$ keep the
templated phrasing, and mean grounding over all queries is $0.995$.

\begin{table}[t]
\caption{The twelve synthetic domains, native operation counts, and the
intervention-training split used in RQ4. Each domain defines a grouped native tool schema (system context, callable
operations, and typed arguments); the full specification is released with the
artifact.}
\label{tab:synthetic_domains}
\begin{center}
\small
\begin{tabular}{@{}llrc@{}}
\toprule
\textbf{Domain} & \textbf{Representative operations} & \textbf{Ops} & \textbf{RQ4} \\
\midrule
Car cabin controller & windows, sunroof, seats, cabin climate & 18 & train \\
Home HVAC / thermostat & modes, zones, holds, fan settings & 13 & train \\
Multi-room media player & playback, sources, queues, rooms & 15 & train \\
Smart kitchen appliances & oven, stove, timers, presets & 13 & train \\
Smart laundry & wash/dry cycles, soil and water levels & 13 & train \\
Home security & locks, alarms, sensors, access codes & 13 & train \\
Smartwatch \& fitness & workouts, goals, hydration, notifications & 13 & train \\
Car navigation & routes, waypoints, ETA, traffic & 13 & train \\
\midrule
Calendar \& scheduling & events, invites, reminders, availability & 15 & held out \\
Smart home controller & lights, scenes, rooms, automations & 16 & held out \\
Smart TV & inputs, apps, search, playback & 13 & held out \\
Robot vacuum & cleaning runs, maps, rooms, schedules & 13 & held out \\
\midrule
\multicolumn{2}{l}{\textbf{Total}} & \textbf{168} & \\
\bottomrule
\end{tabular}
\end{center}
\end{table}

\begin{table}[h]
\caption{Statistics of the controlled synthetic environment: the native tool
catalog (top) and the 2{,}748 evaluation queries (bottom). Every query is a
single user utterance whose target is one to six native calls within one
domain; there are no multi-turn dialogues. Word counts are for the colloquial
queries used in all experiments, with the templated source query in
parentheses.}
\label{tab:synthetic_stats}
\begin{center}
\small
\begin{tabular}{@{}lr@{}}
\toprule
\textbf{Native catalog} & \\
\midrule
Domains & 12 \\
Native tools & 168 (13--18 per domain) \\
Arguments & 397 (1--4 per tool, mean 2.4; all required) \\
\quad enumerated (string enum) & 261 (2--19 values, mean 3.8) \\
\quad free string / integer / boolean & 43 / 92 / 1 \\
\midrule
\textbf{Queries} & \\
\midrule
Queries & 2{,}748 (198--265 per domain) \\
Gold native calls & 7{,}818 (1--6 per query, mean 2.8) \\
\quad single: one call & 585 \\
\quad compound: 2 / 3 / 4 / 5 / 6 independent calls & 441 / 599 / 503 / 202 / 2 \\
\quad compositional: 3 / 4 dependent calls & 144 / 272 \\
Tools covered by gold calls & 168 of 168 (18--97 calls per tool, median 47) \\
Arguments per gold call & 2.4 \\
Query length, words & 11--149, median 80 (templated: 9--31 mean) \\
\quad single / compound / compositional, mean & 66 / 84 / 88 (9 / 28 / 31) \\
\bottomrule
\end{tabular}
\end{center}
\end{table}

\subsection{Real benchmarks}
\label{sec:app-benchmarks}

The real-benchmark runs (Appendix~\ref{sec:app-real-proxy}) use a schema proxy,
so that every benchmark is rewritten by the same code. The proxy is an
OpenAI-compatible server between the harness and the model. It shows the model
the variant schema, decodes every call back to the native tool, and runs the
native call in the benchmark's own environment. It allows no self-correction.
A call that the variant schema rejects is dropped, with no error message and no retry, and a
turn left without any valid call ends as text. The multi-step protocols
(transaction, schema discovery) answer their own intermediate steps, for at most
eight model calls per turn. The model's history shows its own variant calls
rather than the decoded native ones: with native names in its history, the model
copies them on the next turn and those calls are dropped. On \tautwo{}, the
model under test is also the user simulator and the natural-language judge. The benchmarks and the
task sets are \tautwo{} airline (50 tasks) and retail (114), BFCL v3 multi-turn
base (200), AutomationBench with its limited Zapier toolset (599 of 600 tasks;
one task names a tool the toolset lacks), and MCP-Atlas (the 89 tasks whose
reference trajectories use only tools of servers that run without API keys).
Each benchmark keeps its own scorer. MCP-Atlas is scored as mean claim coverage
by a Qwen3.8-27B judge.

\subsection{Models}
\label{sec:app-models}

Nine open-weight models are evaluated on all 32 variants: seven
\emph{earlier-generation} models from three families, from 4B to 32B parameters
(Qwen3-4B and Qwen3-30B-A3B, \citealp{qwen2025qwen3}; Qwen2.5-7B/14B/32B,
\citealp{qwen2024qwen25}; Llama-3.1-8B, \citealp{dubey2024llama}; and
Gemma-4-12B, \citealp{gemma2026gemma4}), and two \emph{newer-generation} Qwen3.5
models (27B and 35B-A3B; \citealp{qwen2026qwen35}), which use that
generation's native tool-call format with thinking enabled. Qwen3.5-4B and 9B
additionally appear on the merge argument structures
(Table~\ref{tab:merge_enc}). Two closed models are evaluated through their APIs
on the thirteen representative variants of Figure~\ref{fig:landscape}, with the
same queries and scorer: gpt-6-luna \citep{openai2026gpt6} (OpenAI Responses API,
reasoning effort medium) and GLM-5.3-Flash \citep{glm2026glm5} (default settings),
one run each in September 2026. Open-weight models are served with vLLM
\citep{kwon2023vllm}.

\subsection{Evaluation protocol}
\label{sec:app-protocol}

Every model and schema variant is evaluated on the same 2{,}748 queries. An
episode succeeds when the multiset of native actions it executes equals the
gold multiset exactly, so the scoring rule does not depend on the schema used to
express the actions. The environment rejects tool calls that violate the active
schema and returns a recoverable error, so the agent can retry after reading the
feedback. Unless noted, results use this state-based exact scoring with invalid
calls rejected, and deterministic decoding.
\FloatBarrier

\section{Schema variants}
\label{sec:app-variants-all}

\subsection{Methods of each operator}
\label{sec:app-methods}

Section~\ref{sec:axes} names only the representative method of each operator
(Table~\ref{tab:axes}). This section lists every method evaluated in the paper;
Appendix~\ref{sec:app-schemas} shows an excerpt of each, and
Appendix~\ref{sec:app-variants} reports their scores.

\textbf{Merge: which functions go together.} \emph{Fully merged} puts all
functions of a domain into one merged tool, which selects the function through an \texttt{operation} argument. \emph{Class dispatch} puts the
functions of each class, across the whole 168-function catalog, into one
merged tool per class (twelve merged tools). \emph{Grouped merges} split a
domain into merged tools of about two, four, or eight functions each, formed by
semantic similarity, at random, or by deliberately mixing unrelated functions.
\emph{Partial merges} fuse only part of a domain and form the merge side of the
granularity ladder (Figure~\ref{fig:rq1_granularity}); they are labelled by the
resulting function count relative to native.

\textbf{Merge: how the merged tool takes its arguments.} All three argument
structures decode to the same native call; they differ only in where the model
states which function it means and where it places that function's arguments.
In the \emph{flat} structure, the merged tool has an \texttt{operation} enum and
one optional field for every argument of every member function, prefixed by the
function name (\texttt{set\_thermostat::zone}). In the \emph{nested} structure,
it has the \texttt{operation} enum and a single \texttt{arguments} object whose
properties are the union of the member functions' arguments, the shape used by
JSON-RPC \citep{jsonrpc2010} and MCP \citep{anthropic2024mcp}. In the \emph{union} structure, there is no \texttt{operation}
field: the merged tool has one argument object per member function, and the model
selects the function by filling exactly one of them. Averaged over domains, the
fully merged flat, nested, and union schemas have about 1{,}300, 830, and
1{,}815 tokens and 34, 2, and 14 top-level arguments. JSON-Schema \texttt{oneOf}
is not used because tool-call parsers do not support it reliably, which would
confound parser and model failures.

\textbf{Split.} \emph{Fully split} bakes every enumerable argument into the tool
name, so \texttt{set\_mute(zone, state)} becomes
\texttt{set\_mute\_\_state\_muted(zone)} and
\texttt{set\_mute\_\_state\_unmuted(zone)}; the split side of the granularity
ladder bakes in progressively more arguments, up to $10.6\times$ the native
function count. \emph{Interval split} does the same for one numeric argument by
cutting its range into two, three, or five intervals (one, two, or four cut
points).

\textbf{Per-tool representation.} \emph{Nest} packs a subset of a tool's
arguments into one nested object (one level deep in all our variants).
\emph{Rename} either prefixes tool names with their class as a namespace
(\emph{namespaced names}) or replaces them with opaque strings (\emph{opaque
names}). The \emph{whole catalog} control is not a rename: it exposes all 168
functions with their native names and serves as the control for class dispatch
and namespaced names, which expose the same catalog. \emph{Strip descriptions}
and \emph{reorder arguments} have a single method each.

\textbf{Cross-call protocols.} \emph{Transaction} applies the
open--write--confirm sequence either to every function or to a subset, and can
be composed with reference resolution so that the written arguments are handles
rather than values. \emph{Reference resolution} and \emph{schema discovery} have
a single method each.

\subsection{Excerpts of each method}
\label{sec:app-schemas}

Each listing is the climate-domain tool \texttt{set\_thermostat} unless noted.
Blue marks what the operator changes relative to native; gray \texttt{...}
omits unchanged fields or sibling tools. Flatten-arguments is omitted because
the native parameters are already flat.

\paragraph{Native.}
Default original tool schema.
\begin{lstlisting}[style=toolschema]
{
  "name": "set_thermostat",
  "description": "Set a zone thermostat mode and target temperature.",
  "parameters": {
    "zone": {"type": "string", "enum": ["upstairs", "downstairs", "basement", "whole_house"]},
    "mode": {"type": "string", "enum": ["heat", "cool", "auto", "eco", "off"]},
    "temperature": {"type": "integer"}
  },
  "required": ["zone", "mode", "temperature"]
}
\end{lstlisting}

\paragraph{Merge, flat argument structure.}
Thirteen named tools become one merged tool. Every member's arguments appear
as optional namespaced fields on that one tool; hard validation rejects
arguments that do not belong to the chosen \texttt{operation}.
\begin{lstlisting}[style=toolschema]
{
  "name": "(*@\diff{execute}@*)",
  "parameters": {
    "(*@\diff{operation}@*)": {"type": "string", "enum": ["set_thermostat", "set_temp_range", (*@\omitrest@*)]},
    "(*@\diff{set\_thermostat::zone}@*)": {"type": "string", "enum": ["upstairs", (*@\omitrest@*)], "optional": (*@\diff{true}@*)},
    "(*@\diff{set\_thermostat::mode}@*)": {"type": "string", "enum": ["heat", (*@\omitrest@*)], "optional": (*@\diff{true}@*)},
    "(*@\diff{set\_thermostat::temperature}@*)": {"type": "integer", "optional": (*@\diff{true}@*)},
    "(*@\diff{set\_temp\_range::heat\_to}@*)": {"type": "integer", "optional": (*@\diff{true}@*)},
    (*@\omitrest@*)
  },
  "required": ["(*@\diff{operation}@*)"]
}
\end{lstlisting}

\paragraph{Merge, nested argument structure.}
Same members and \texttt{operation} enum. Arguments sit in one shared object
whose properties are the \emph{union} of member parameters, not a per-operation
copy.
\begin{lstlisting}[style=toolschema]
{
  "name": "execute",
  "parameters": {
    "operation": {"type": "string", "enum": ["set_thermostat", (*@\omitrest@*)]},
    "(*@\diff{arguments}@*)": {
      "zone": {"type": "string", "enum": ["upstairs", (*@\omitrest@*)]},
      "mode": {"type": "string", "enum": ["heat", (*@\omitrest@*)]},
      "temperature": {"type": "integer"},
      "heat_to": {"type": "integer"},
      (*@\omitrest@*)
    }
  }
}
\end{lstlisting}

\paragraph{Merge, union argument structure.}
There is no \texttt{operation} enum. The agent selects an operation by which
object is present and leaves the others unset.
\begin{lstlisting}[style=toolschema]
{
  "name": "execute",
  "parameters": {
    "(*@\diff{set\_thermostat}@*)": {
      "zone": {"type": "string", "enum": ["upstairs", (*@\omitrest@*)]},
      "mode": {"type": "string", "enum": ["heat", (*@\omitrest@*)]},
      "temperature": {"type": "integer"}
    },
    "(*@\diff{set\_temp\_range}@*)": {
      "zone": {"type": "string"},
      "heat_to": {"type": "integer"},
      "cool_to": {"type": "integer"}
    },
    (*@\omitrest@*)
  }
}
\end{lstlisting}

\paragraph{Merge, class dispatch (catalog scope, class grouping, flat).}
Same flat argument structure, one merged tool per class, named by the class. On
this single-class catalog that is one tool; a 12-class catalog has 12.
\begin{lstlisting}[style=toolschema]
{
  "name": "(*@\diff{climate}@*)",
  "parameters": {
    "operation": {"type": "string", "enum": ["set_thermostat", "set_temp_range", (*@\omitrest@*)]},
    "set_thermostat::zone": {"type": "string", "optional": true},
    (*@\omitrest@*)
  },
  "required": ["operation"]
}
\end{lstlisting}

\paragraph{Merge, semantic grouping.}
This is still a flat merge. Similar names and parameter sets are clustered,
then each cluster is merged, so climate's 13 tools become three merged tools
rather than one. The argument structure matches merge-flat (an \texttt{operation} enum plus
namespaced optional arguments). Random and anti-semantic grouping keep these
three group sizes and scramble membership.
\begin{lstlisting}[style=toolschema]
{
  "name": "(*@\diff{thermostat\_humidity\_ops}@*)",
  "parameters": {
    "operation": {"enum": ["set_thermostat", "set_thermostat_hold",
      "set_temp_range", "set_humidity", "set_humidity_alert",
      "set_dehumidifier", "set_air_purifier", "set_vent", "set_zone_priority"]},
    "set_thermostat::zone": {"type": "string", "optional": true}
  }
}
{
  "name": "(*@\diff{mode\_fan\_ops}@*)",
  "parameters": {
    "operation": {"enum": ["set_fan_mode", "set_away_mode", "set_hvac_schedule"]}
  }
}
{
  "name": "(*@\diff{filter\_reminder\_ops}@*)",
  "parameters": {
    "operation": {"enum": ["set_filter_reminder"]}
  }
}
\end{lstlisting}

\paragraph{Split by enum value (fully split).}
\texttt{mode} is removed as an argument and baked into the function name.
The catalog replaces one tool with five; two are shown.
\begin{lstlisting}[style=toolschema]
{
  "name": "(*@\diff{set\_thermostat\_\_mode\_heat}@*)",
  "parameters": {
    "zone": {"type": "string", "enum": ["upstairs", (*@\omitrest@*)]},
    "temperature": {"type": "integer"}
  },
  "required": ["zone", "temperature"]
}
{
  "name": "(*@\diff{set\_thermostat\_\_mode\_cool}@*)",
  "parameters": { (*@\omitrest@*) }
}
\end{lstlisting}

\paragraph{Split by numeric predicate (interval split).}
\texttt{set\_thermostat} becomes two tools that still take \texttt{temperature},
but each accepts only one side of the cut.
\begin{lstlisting}[style=toolschema]
{
  "name": "(*@\diff{set\_thermostat\_\_temperature\_low}@*)",
  "parameters": {
    "zone": {"type": "string", "enum": ["upstairs", (*@\omitrest@*)]},
    "mode": {"type": "string", "enum": ["heat", (*@\omitrest@*)]},
    "temperature": {"type": "integer", "(*@\diff{exclusiveMaximum}@*)": (*@\diff{20}@*)}
  }
}
{
  "name": "(*@\diff{set\_thermostat\_\_temperature\_high}@*)",
  "parameters": {
    "zone": {"type": "string", "enum": ["upstairs", (*@\omitrest@*)]},
    "mode": {"type": "string", "enum": ["heat", (*@\omitrest@*)]},
    "temperature": {"type": "integer", "(*@\diff{minimum}@*)": (*@\diff{20}@*)}
  }
}
\end{lstlisting}

\paragraph{Nest arguments.}
The native three-argument object becomes one required wrapper. The inner
fields are unchanged.
\begin{lstlisting}[style=toolschema]
{
  "name": "set_thermostat",
  "parameters": {
    "(*@\diff{options}@*)": {
      "zone": {"type": "string", "enum": ["upstairs", (*@\omitrest@*)]},
      "mode": {"type": "string", "enum": ["heat", (*@\omitrest@*)]},
      "temperature": {"type": "integer"}
    }
  },
  "required": ["(*@\diff{options}@*)"]
}
\end{lstlisting}

\paragraph{Rename, namespaced names.}
Only the function name changes: the class name is prefixed as a namespace;
arguments stay native.
\begin{lstlisting}[style=toolschema]
{
  "name": "(*@\diff{climate\_\_set\_thermostat}@*)",
  "parameters": {
    "zone": {"type": "string", "enum": ["upstairs", (*@\omitrest@*)]},
    "mode": {"type": "string", "enum": ["heat", (*@\omitrest@*)]},
    "temperature": {"type": "integer"}
  }
}
\end{lstlisting}

\paragraph{Rename, opaque names.}
The name becomes an opaque identifier; the description and arguments stay
native, so the description is the remaining routing signal.
\begin{lstlisting}[style=toolschema]
{
  "name": "(*@\diff{fn\_1748c5}@*)",
  "description": "Set a zone thermostat mode and target temperature.",
  "parameters": {
    "zone": {"type": "string", "enum": ["upstairs", (*@\omitrest@*)]},
    "mode": {"type": "string", "enum": ["heat", (*@\omitrest@*)]},
    "temperature": {"type": "integer"}
  }
}
\end{lstlisting}

\paragraph{Strip descriptions.}
Function and parameter descriptions are removed; names and types stay.
\begin{lstlisting}[style=toolschema]
{
  "name": "set_thermostat",
  "description": (*@\diff{""}@*),
  "parameters": {
    "zone": {"type": "string", "enum": ["upstairs", "downstairs", "basement", "whole_house"]},
    "mode": {"type": "string", "enum": ["heat", "cool", "auto", "eco", "off"]},
    "temperature": {"type": "integer"}
  }
}
\end{lstlisting}

\paragraph{Reorder arguments.}
Names and types are unchanged. The declared property order, and therefore
\texttt{required}, becomes \texttt{temperature}, \texttt{mode}, \texttt{zone}.
\begin{lstlisting}[style=toolschema]
{
  "name": "set_thermostat",
  "parameters": {
    "temperature": {"type": "integer"},
    "mode": {"type": "string", "enum": ["heat", (*@\omitrest@*)]},
    "zone": {"type": "string", "enum": ["upstairs", (*@\omitrest@*)]}
  },
  "required": ["(*@\diff{temperature}@*)", "(*@\diff{mode}@*)", "(*@\diff{zone}@*)"]
}
\end{lstlisting}

\paragraph{Transaction.}
One native call becomes five dependent calls. Nothing is executed until
confirm. The \texttt{mode} and \texttt{temperature} writes match the
\texttt{zone} write.
\begin{lstlisting}[style=toolschema]
{
  "name": "(*@\diff{begin\_set\_thermostat}@*)",
  "parameters": {}
}
{
  "name": "(*@\diff{set\_set\_thermostat\_\_zone}@*)",
  "parameters": {
    "(*@\diff{txn\_id}@*)": {"type": "string"},
    "zone": {"type": "string", "enum": ["upstairs", (*@\omitrest@*)]}
  },
  "required": ["(*@\diff{txn\_id}@*)", "zone"]
}
{
  "name": "(*@\diff{set\_set\_thermostat\_\_mode}@*)",
  "parameters": { "txn_id": {"type": "string"}, "mode": {"type": "string", (*@\omitrest@*)} }
}
{
  "name": "(*@\diff{set\_set\_thermostat\_\_temperature}@*)",
  "parameters": { "txn_id": {"type": "string"}, "temperature": {"type": "integer"} }
}
{
  "name": "(*@\diff{commit\_set\_thermostat}@*)",
  "parameters": {
    "(*@\diff{txn\_id}@*)": {"type": "string"}
  },
  "required": ["(*@\diff{txn\_id}@*)"]
}
\end{lstlisting}

\paragraph{Reference resolution.}
\texttt{zone} is removed from \texttt{set\_thermostat}. The agent must first
call a resolver with a free-form query and then pass the returned handle.
\begin{lstlisting}[style=toolschema]
{
  "name": "(*@\diff{resolve\_set\_thermostat\_\_zone}@*)",
  "parameters": {
    "(*@\diff{query}@*)": {"type": "string"}
  },
  "required": ["(*@\diff{query}@*)"]
}
{
  "name": "set_thermostat",
  "parameters": {
    "mode": {"type": "string", "enum": ["heat", (*@\omitrest@*)]},
    "temperature": {"type": "integer"},
    "(*@\diff{zone\_ref}@*)": {"type": "string"}
  },
  "required": ["(*@\diff{zone\_ref}@*)", "mode", "temperature"]
}
\end{lstlisting}

\paragraph{Schema discovery.}
The upfront catalog has no \texttt{set\_thermostat} and no argument types.
\texttt{list\_methods} returns those native parameter schemas at runtime;
\texttt{invoke} accepts an untyped \texttt{arguments} object.
\begin{lstlisting}[style=toolschema]
{
  "name": "(*@\diff{list\_methods}@*)",
  "parameters": {
    "class": {"type": "string", "enum": ["climate"]}
  }
}
{
  "name": "(*@\diff{invoke}@*)",
  "parameters": {
    "class": {"type": "string", "enum": ["climate"]},
    "method": {"type": "string"},
    "arguments": {"type": "object"}
  }
}
\end{lstlisting}
Runtime result of \texttt{list\_methods(\{"class": "climate"\})}, abbreviated:
\begin{lstlisting}[style=toolschema]
{
  "methods": [
    {
      "name": "set_thermostat",
      "parameters": {
        "zone": {"type": "string", "enum": ["upstairs", (*@\omitrest@*)]},
        "mode": {"type": "string", "enum": ["heat", (*@\omitrest@*)]},
        "temperature": {"type": "integer"}
      }
    },
    (*@\omitrest@*)
  ]
}
\end{lstlisting}

\subsection{Scores of all methods}
\label{sec:app-variants}

\begin{figure*}[t]
\centering
\includegraphics[width=\textwidth]{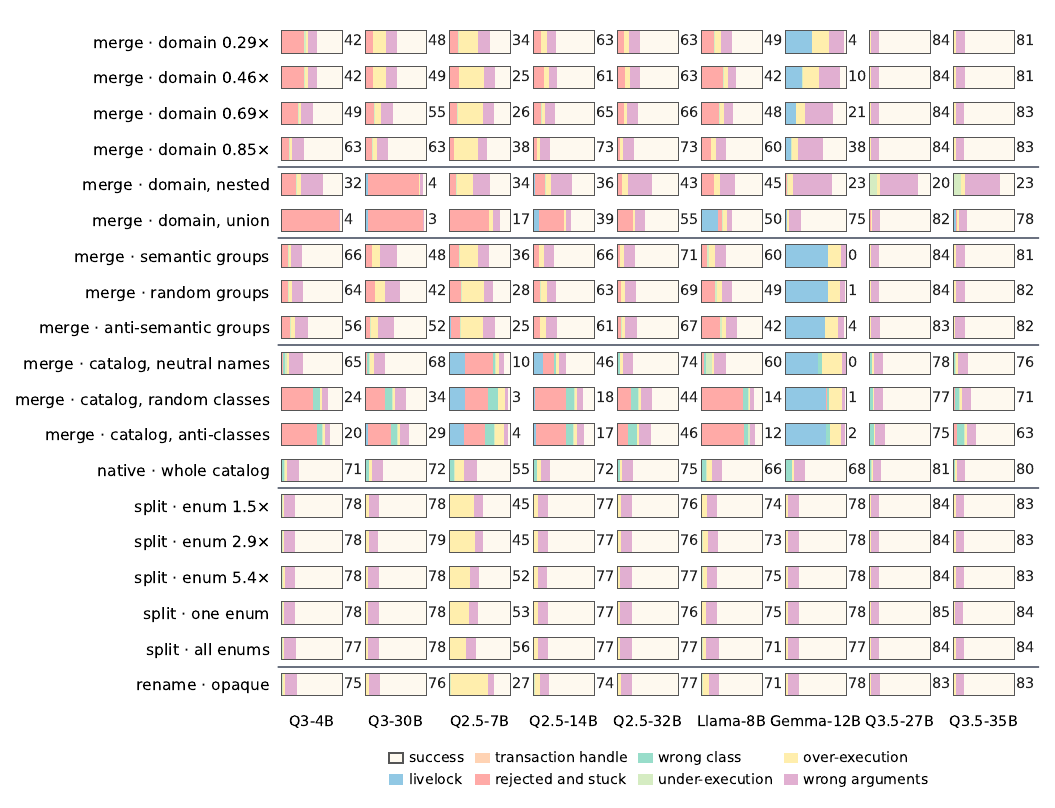}  
\caption{\textbf{Scores of the remaining methods of each operator}, in the
format of Figure~\ref{fig:landscape} (nine models, 2{,}748 queries). Row labels
give the operator and method (Table~\ref{tab:axes}): partial merges along the
granularity ladder (function count relative to native), the nested and union
argument structures of the fully merged schema, the three grouping rules at a
fixed group size, the membership and naming controls of class dispatch together
with the whole-catalog control, the split ladder, and opaque names.}
\label{fig:app-variants}
\end{figure*}

Figure~\ref{fig:landscape} shows one representative method per operator.
Figure~\ref{fig:app-variants} reports the remaining methods in the same format
on all nine models, and Table~\ref{tab:merge_enc} gives the fully merged schema
under each argument structure, including two further newer-generation models
(Qwen3.5-4B and 9B). Within an operator, the score moves with the method about
as much as it moves across operators, which is why the main figures fix one
method per operator and the tables below vary one factor at a time.

\FloatBarrier

\section{Additional results on measuring schema bias (RQ1)}
\label{sec:app-rq1}

\subsection{Argument structures and mixed calling conventions}
\label{sec:app-mixtures}

\textbf{Argument structure of a merged tool.} Table~\ref{tab:merge_enc} shows
that no argument structure is safe for every model: nested is the worst
structure for most models and costs every newer-generation model 58 to 64
points, while flat is Gemma-4-12B's worst structure (it scores zero) and union
is worst for the two Qwen3 models and Qwen2.5-7B. The same ordering holds for
the semantic grouped merge (about 3.3 merged tools per domain): nested costs 29
to 60 points on every model, union 0 to 32, and flat under 30 for every model
except Gemma-4-12B (78). Under the union structure, Qwen3.5-9B produced
malformed tool calls on $24.8\%$ of requests; these are scored as model
failures.

\begin{table}[t]
\caption{Argument structure of merged tools, from full merging to smaller semantic groups. \emph{Fully merged}
puts each domain into one tool (the member sets of the fully merged row in Figure~\ref{fig:landscape}), about
$0.07\times$ the native function count; \emph{semantic groups} merge each domain into groups of related functions
at about $0.3\times$ or $0.5\times$ the native function count, with the same groups, names, and function count in
every column of a block. A merged tool takes its arguments in one of three structures: \emph{flat} (an
\texttt{operation} enum plus every member's arguments as optional prefixed fields), \emph{nested} (the
\texttt{operation} enum plus one \texttt{arguments} object), or \emph{union} (one argument object per member,
no \texttt{operation} field); \emph{mixed} rotates the three structures across the merged tools of a catalog
(average of three balanced rotations). \emph{Native} is the model's native success; every other cell is the change
from native, except \emph{mixed $-$ avg.}: mixed minus the average of the three single-structure catalogs of the
same block, which is near zero if mixing is harmless and negative if mixing costs more than its parts. Exact success on 2{,}748 tasks; shading as in Table~\ref{tab:rq4-methods}; $^{\circ}$ marks a mixed $-$ avg.\ value
whose paired-bootstrap 95\% interval includes zero; -- = not run.}
\label{tab:merge_enc}
\begin{center}
\scriptsize
\setlength{\tabcolsep}{2.2pt}
\resizebox{\linewidth}{!}{%
\begin{tabular}{@{}l c ccc ccccc ccccc@{}}
\toprule
& & \multicolumn{3}{c}{merge $\cdot$ domain} & \multicolumn{5}{c}{merge $\cdot$ semantic groups, $0.3\times$} & \multicolumn{5}{c}{merge $\cdot$ semantic groups, $0.5\times$} \\
\cmidrule(lr){3-5} \cmidrule(lr){6-10} \cmidrule(l){11-15}
Model & Native & Flat & Nested & Union & Flat & Nested & Union & Mixed & \shortstack{Mixed\\$-$ avg.} & Flat & Nested & Union & Mixed & \shortstack{Mixed\\$-$ avg.} \\
\midrule
Qwen3-4B & 0.77 & \cellcolor{deltaloss!83}$-$0.41 & \cellcolor{deltaloss!90}$-$0.45 & \cellcolor{deltaloss!90}$-$0.74 & \cellcolor{deltaloss!42}$-$0.16 & \cellcolor{deltaloss!80}$-$0.39 & \cellcolor{deltaloss!47}$-$0.19 & \cellcolor{deltaloss!76}$-$0.36 & \cellcolor{deltaloss!34}$-$0.12 & \cellcolor{deltaloss!44}$-$0.17 & \cellcolor{deltaloss!74}$-$0.36 & \cellcolor{deltaloss!51}$-$0.22 & \cellcolor{deltaloss!85}$-$0.42 & \cellcolor{deltaloss!44}$-$0.17 \\
Qwen3-30B-A3B & 0.77 & \cellcolor{deltaloss!90}$-$0.70 & \cellcolor{deltaloss!90}$-$0.73 & \cellcolor{deltaloss!90}$-$0.74 & \cellcolor{deltaloss!40}$-$0.15 & \cellcolor{deltaloss!90}$-$0.59 & \cellcolor{deltaloss!40}$-$0.15 & \cellcolor{deltaloss!90}$-$0.48 & \cellcolor{deltaloss!45}$-$0.18 & \cellcolor{deltaloss!41}$-$0.16 & \cellcolor{deltaloss!90}$-$0.56 & \cellcolor{deltaloss!47}$-$0.19 & \cellcolor{deltaloss!90}$-$0.53 & \cellcolor{deltaloss!54}$-$0.23 \\
Qwen2.5-7B & 0.53 & \cellcolor{deltaloss!44}$-$0.17 & \cellcolor{deltaloss!47}$-$0.19 & \cellcolor{deltaloss!74}$-$0.36 & \cellcolor{deltaloss!48}$-$0.20 & \cellcolor{deltaloss!81}$-$0.40 & \cellcolor{deltaloss!52}$-$0.22 & \cellcolor{deltaloss!70}$-$0.33 & \cellcolor{deltaloss!24}$-$0.06 & \cellcolor{deltaloss!48}$-$0.20 & \cellcolor{deltaloss!87}$-$0.43 & \cellcolor{deltaloss!51}$-$0.22 & \cellcolor{deltaloss!80}$-$0.39 & \cellcolor{deltaloss!32}$-$0.10 \\
Qwen2.5-14B & 0.77 & \cellcolor{deltaloss!52}$-$0.22 & \cellcolor{deltaloss!84}$-$0.41 & \cellcolor{deltaloss!79}$-$0.39 & \cellcolor{deltaloss!37}$-$0.13 & \cellcolor{deltaloss!73}$-$0.35 & \cellcolor{deltaloss!42}$-$0.16 & \cellcolor{deltaloss!51}$-$0.22 & 0.00$^{\circ}$ & \cellcolor{deltaloss!34}$-$0.12 & \cellcolor{deltaloss!66}$-$0.30 & \cellcolor{deltaloss!43}$-$0.17 & \cellcolor{deltaloss!47}$-$0.19 & +0.01$^{\circ}$ \\
Qwen2.5-32B & 0.76 & $-$0.01 & \cellcolor{deltaloss!70}$-$0.33 & \cellcolor{deltaloss!51}$-$0.22 & \cellcolor{deltaloss!26}$-$0.07 & \cellcolor{deltaloss!63}$-$0.29 & \cellcolor{deltaloss!27}$-$0.07 & \cellcolor{deltaloss!42}$-$0.17 & $-$0.02 & \cellcolor{deltaloss!23}$-$0.05 & \cellcolor{deltaloss!60}$-$0.27 & \cellcolor{deltaloss!22}$-$0.04 & \cellcolor{deltaloss!38}$-$0.14 & $-$0.02 \\
Llama-3.1-8B & 0.74 & \cellcolor{deltaloss!24}$-$0.05 & \cellcolor{deltaloss!62}$-$0.28 & \cellcolor{deltaloss!55}$-$0.24 & \cellcolor{deltaloss!45}$-$0.18 & \cellcolor{deltaloss!67}$-$0.31 & \cellcolor{deltaloss!55}$-$0.24 & \cellcolor{deltaloss!61}$-$0.27 & \cellcolor{deltaloss!20}$-$0.03 & \cellcolor{deltaloss!40}$-$0.15 & \cellcolor{deltaloss!56}$-$0.24 & \cellcolor{deltaloss!42}$-$0.17 & \cellcolor{deltaloss!57}$-$0.26 & \cellcolor{deltaloss!26}$-$0.07 \\
Gemma-4-12B & 0.79 & \cellcolor{deltaloss!90}$-$0.79 & \cellcolor{deltaloss!90}$-$0.55 & \cellcolor{deltaloss!21}$-$0.03 & \cellcolor{deltaloss!90}$-$0.75 & \cellcolor{deltaloss!84}$-$0.41 & $-$0.01 & \cellcolor{deltaloss!90}$-$0.52 & \cellcolor{deltaloss!36}$-$0.13 & \cellcolor{deltaloss!90}$-$0.76 & \cellcolor{deltaloss!68}$-$0.32 & 0.00 & \cellcolor{deltaloss!90}$-$0.54 & \cellcolor{deltaloss!45}$-$0.18 \\
\midrule
Qwen3.5-4B & 0.83 & $-$0.03 & \cellcolor{deltaloss!90}$-$0.58 & \cellcolor{deltaloss!34}$-$0.12 & -- & -- & -- & -- & -- & -- & -- & -- & -- & -- \\
Qwen3.5-9B & 0.83 & \cellcolor{deltaloss!26}$-$0.06 & \cellcolor{deltaloss!90}$-$0.60 & \cellcolor{deltaloss!90}$-$0.47 & -- & -- & -- & -- & -- & -- & -- & -- & -- & -- \\
Qwen3.5-27B & 0.84 & $-$0.01 & \cellcolor{deltaloss!90}$-$0.64 & $-$0.02 & $-$0.01 & \cellcolor{deltaloss!90}$-$0.55 & 0.00 & \cellcolor{deltaloss!54}$-$0.24 & \cellcolor{deltaloss!23}$-$0.05 & $-$0.01 & \cellcolor{deltaloss!85}$-$0.42 & $-$0.01 & \cellcolor{deltaloss!45}$-$0.18 & \cellcolor{deltaloss!21}$-$0.03 \\
Qwen3.5-35B-A3B & 0.83 & \cellcolor{deltaloss!22}$-$0.04 & \cellcolor{deltaloss!90}$-$0.61 & \cellcolor{deltaloss!24}$-$0.06 & $-$0.02 & \cellcolor{deltaloss!90}$-$0.50 & $-$0.01 & \cellcolor{deltaloss!53}$-$0.23 & \cellcolor{deltaloss!24}$-$0.05 & $-$0.01 & \cellcolor{deltaloss!80}$-$0.39 & $-$0.02 & \cellcolor{deltaloss!49}$-$0.20 & \cellcolor{deltaloss!25}$-$0.06 \\
\bottomrule
\end{tabular}}
\end{center}
\end{table}

\textbf{Mixing argument structures in one catalog.} In the granularity ladder,
partial merges, which mix one merged tool with native named tools, score below
both endpoints for Llama and Qwen2.5 (Llama: $0.71$, $0.42$, and $0.75$ for
fully merged, partial, and native), and the same dip appears on \tautwo{}
airline. Partial merges, however, change the function count and the merged
membership at the same time as they mix conventions. The semantic-group blocks of
Table~\ref{tab:merge_enc} isolate the mixing: at a fixed partition into
merged tools (about $0.3\times$ and $0.5\times$ the native function count),
every catalog exposes the same groups, names, and function count and differs
only in whether all merged tools use one argument structure or the three
structures rotate across groups. The mixed catalogs have between 1{,}618 and
1{,}904 tokens, inside the range of the homogeneous ones (1{,}349 to 2{,}137).
Mixing costs more than the average of its parts: mixed $-$ avg.\ is negative with an
interval excluding zero in 16 of the 18 model-by-granularity cells, and the two
exceptions (Qwen2.5-14B) are within one point of zero. A mixture is nevertheless
almost always better than its worst structure used throughout; the only
significant exception is Qwen3-4B at $0.5\times$ ($-0.065$ $[-0.080,-0.048]$). Mixing
protocols across calls behaves differently: a catalog in which half the
functions use the transaction protocol interpolates between native and fully
transactional ($0.36$, within $[0.28, 0.77]$, on Qwen3-4B).

\subsection{Which operations share a merged tool}
\label{sec:app-grouping}

\textbf{Within a domain.} Three grouped merges with identical group count and
sizes, differing only in which operations share a merged tool, follow the order
semantic, then random, then anti-semantic for five of the seven
earlier-generation models, with a semantic-minus-anti-semantic difference of
$3.9$ (Qwen2.5-32B) to $18.2$ points (Llama-3.1-8B). Two exceptions qualify the
pattern: all three groupings score near zero for Gemma-4-12B, and Qwen3-30B-A3B
reverses the order (anti-semantic $0.52$, semantic $0.48$). The two
newer-generation models stay within one point across the three groupings. On
Qwen3-4B, semantic minus anti-semantic grows with group size ($3.8$, $9.3$, and
$12.6$ points at two, four, and eight functions per merged tool): semantic
grouping improves as groups grow, until everything is merged into one
tool, while anti-semantic grouping steadily worsens. Short of full
merging, most of the loss attributed to coarser grouping is explained by which
operations are grouped together.

\textbf{Across the whole catalog: membership, not naming.} When all 168
functions are exposed at once, plain and namespaced names keep one tool per
function and score nearly identically ($0.71$ and $0.70$ on Qwen3-4B), so the
gap between namespaced names and native in Figure~\ref{fig:landscape} is the
cost of the larger catalog, not of the prefix. Class dispatch (about 14
functions in each of 12 class-named merged tools) scores $0.66$, far above one
fully merged tool ($0.36$), but that comparison changes the number of
merged tools, their membership, and their names at once.
Table~\ref{tab:class_group} holds the flat structure, twelve merged tools, and
the group-size profile fixed and varies only membership and naming: the class
groups with class names, the same groups with neutral names
(\texttt{group\_01}--\texttt{group\_12}, with descriptions of matched length
but no class information), and five random and five anti-class partitions
(fixed seeds; anti-class partitions spread each class across merged tools as
evenly as possible), also with neutral names. Membership matters: class groups
beat random groups by $+6$ (Qwen2.5-7B) to $+46$ points (Llama-3.1-8B), with
intervals above zero for seven of eight models. Naming matters little: at most
$+10$ points (Qwen3-8B), indistinguishable from zero for Qwen3-4B, and slightly
negative for Qwen3-30B-A3B and Qwen2.5-32B, the two models whose dominant merge
failure is calling the wrong merged-tool name. Gemma-4-12B is again the
exception: every arrangement scores below $0.04$ because it fails on any
merged tool with an \texttt{operation} enum.

\begin{table}[t]
\caption{Which functions share a merged tool, across the whole catalog. Every variant exposes all 168 native
functions through twelve merged tools with the flat argument structure and the same group sizes; they differ only
in which functions share a tool and how the tools are named. \emph{Class dispatch} groups each class under the
class name; \emph{neutral names} keeps those groups but names them \texttt{group\_01}--\texttt{group\_12};
\emph{random classes} and \emph{anti-classes} regroup the functions at random or so as to split every class
(mean of five partitions each, neutral names). \emph{Membership effect} is neutral names minus random classes
(does grouping by class help?); \emph{naming effect} is class dispatch minus neutral names (do class names help?).
Exact success on 2{,}748 tasks; shading as in Table~\ref{tab:rq4-methods}; $^{\circ}$ marks an effect whose
paired-bootstrap 95\% interval includes zero. Qwen3-8B, not one of the nine main models, appears in this control only.}
\label{tab:class_group}
\begin{center}
\scriptsize
\begin{tabular*}{\linewidth}{@{\extracolsep{\fill}}l cccc cc@{}}
\toprule
& \multicolumn{4}{c}{merge $\cdot$ catalog (twelve merged tools)} & \multicolumn{2}{c}{Effect} \\
\cmidrule(lr){2-5} \cmidrule(l){6-7}
Model & Class dispatch & Neutral names & Random classes & Anti-classes & Membership & Naming \\
\midrule
Qwen3-4B & 0.66 & 0.66 & 0.23 & 0.21 & \cellcolor{deltagain!85}+0.42 & +0.01$^{\circ}$ \\
Qwen3-8B & 0.64 & 0.53 & 0.20 & 0.20 & \cellcolor{deltagain!71}+0.34 & \cellcolor{deltagain!32}+0.10 \\
Qwen3-30B-A3B & 0.67 & 0.68 & 0.28 & 0.26 & \cellcolor{deltagain!82}+0.40 & $-$0.01 \\
Qwen2.5-7B & 0.18 & 0.10 & 0.04 & 0.03 & \cellcolor{deltagain!25}+0.06 & \cellcolor{deltagain!28}+0.08 \\
Qwen2.5-14B & 0.52 & 0.46 & 0.17 & 0.16 & \cellcolor{deltagain!63}+0.29 & \cellcolor{deltagain!26}+0.06 \\
Qwen2.5-32B & 0.73 & 0.74 & 0.44 & 0.44 & \cellcolor{deltagain!65}+0.30 & $-$0.01 \\
Llama-3.1-8B & 0.61 & 0.60 & 0.14 & 0.12 & \cellcolor{deltagain!90}+0.46 & +0.01$^{\circ}$ \\
Gemma-4-12B & 0.04 & 0.01 & 0.02 & 0.02 & $-$0.01 & \cellcolor{deltagain!20}+0.03 \\
\bottomrule
\end{tabular*}
\end{center}
\end{table}

\subsection{Per-tool representation}
\label{sec:app-surface}

Nesting the arguments, removing descriptions, and permuting argument order each
move success by at most 4 points on every one of the nine models
(Figure~\ref{fig:landscape}), and the few larger moves go in both directions
(Qwen2.5-7B gains 1.3 points without descriptions and loses 2.4 with permuted
arguments). Opaque names (Figure~\ref{fig:app-variants}) lower Qwen2.5-7B from
$0.54$ to $0.27$ but change every other model by at most 3 points; within
Qwen2.5, the drop shrinks from 27 points at 7B to 3 at 14B and disappears at
32B. Names and descriptions are redundant routing signals: descriptions are
unnecessary when names are informative (Qwen3-4B scores $0.78$ without them),
and names are unnecessary when descriptions remain (opaque names cost Qwen3-4B
2 points), except for the smallest Qwen2.5 model.

\subsection{Schema bias on four real benchmarks through a schema proxy}
\label{sec:app-real-proxy}
\begin{table}[!t]
\caption{Schema bias on four real benchmarks with no self-correction (Appendix~\ref{sec:app-real-proxy}). First row:
native score; other cells: change from native, shaded as in Table~\ref{tab:rq4-methods}. Scores: mean reward
(\tautwo{}, 50 and 114 tasks), accuracy (BFCL, 200), mean partial credit (AutomationBench, 599), mean claim coverage
(MCP-Atlas, 89, Qwen3.8-27B judge). n/a: interval split on retail (no numeric arguments); reference resolution is not run.}
\label{tab:real_proxy}
\begin{center}
\scriptsize
\setlength{\tabcolsep}{2.2pt}
\resizebox{0.85\linewidth}{!}{%
\begin{tabular}{@{}l ccccc ccccc@{}}
\toprule
& \multicolumn{5}{c}{Qwen3.5-27B} & \multicolumn{5}{c}{Qwen3-4B} \\
\cmidrule(lr){2-6} \cmidrule(l){7-11}
Variant & $\tau^2$ air. & $\tau^2$ ret. & BFCL & AutoB. & MCP-A. & $\tau^2$ air. & $\tau^2$ ret. & BFCL & AutoB. & MCP-A. \\
\midrule
Native & 0.78 & 0.47 & 0.73 & 0.53 & 0.54 & 0.26 & 0.25 & 0.33 & 0.17 & 0.17 \\
\midrule
\multicolumn{11}{@{}l}{\emph{Tool-set partitioning}} \\
Fully merged & \cellcolor{deltaloss!42}$-$0.16 & \cellcolor{deltaloss!65}$-$0.30 & \cellcolor{deltaloss!90}$-$0.67 & \cellcolor{deltaloss!89}$-$0.44 & \cellcolor{deltaloss!58}$-$0.26 & \cellcolor{deltagain!38}+0.14 & \cellcolor{deltaloss!44}$-$0.18 & \cellcolor{deltaloss!70}$-$0.33 & \cellcolor{deltaloss!43}$-$0.17 & \cellcolor{deltaloss!39}$-$0.14 \\
Class dispatch & $-$0.02 & \cellcolor{deltaloss!31}$-$0.10 & \cellcolor{deltaloss!84}$-$0.41 & \cellcolor{deltaloss!63}$-$0.29 & \cellcolor{deltaloss!82}$-$0.40 & 0.00 & \cellcolor{deltaloss!25}$-$0.06 & \cellcolor{deltaloss!66}$-$0.30 & \cellcolor{deltaloss!23}$-$0.05 & \cellcolor{deltaloss!35}$-$0.12 \\
Fully split & 0.00 & +0.03 & $-$0.01 & \cellcolor{deltaloss!27}$-$0.07 & \cellcolor{deltaloss!25}$-$0.06 & $-$0.02 & +0.02 & $-$0.01 & \cellcolor{deltaloss!23}$-$0.05 & \cellcolor{deltaloss!25}$-$0.06 \\
Interval split & $-$0.02 & n/a & \cellcolor{deltaloss!22}$-$0.04 & \cellcolor{deltaloss!54}$-$0.23 & 0.00 & \cellcolor{deltagain!28}+0.08 & n/a & \cellcolor{deltaloss!29}$-$0.09 & \cellcolor{deltaloss!26}$-$0.07 & \cellcolor{deltaloss!23}$-$0.05 \\
\midrule
\multicolumn{11}{@{}l}{\emph{Per-tool representation}} \\
Nested args & \cellcolor{deltagain!22}+0.04 & +0.03 & \cellcolor{deltaloss!22}$-$0.04 & $-$0.01 & \cellcolor{deltaloss!31}$-$0.09 & \cellcolor{deltagain!28}+0.08 & \cellcolor{deltagain!25}+0.06 & \cellcolor{deltaloss!23}$-$0.05 & \cellcolor{deltaloss!24}$-$0.06 & \cellcolor{deltaloss!20}$-$0.03 \\
Namespaced names & \cellcolor{deltagain!28}+0.08 & $-$0.03 & \cellcolor{deltaloss!23}$-$0.05 & $-$0.02 & \cellcolor{deltaloss!24}$-$0.05 & \cellcolor{deltagain!32}+0.10 & $-$0.03 & \cellcolor{deltaloss!53}$-$0.23 & $-$0.01 & \cellcolor{deltaloss!23}$-$0.05 \\
Strip descriptions & +0.02 & \cellcolor{deltagain!24}+0.05 & \cellcolor{deltaloss!53}$-$0.23 & \cellcolor{deltaloss!33}$-$0.11 & \cellcolor{deltaloss!27}$-$0.07 & +0.02 & +0.01 & \cellcolor{deltaloss!31}$-$0.10 & \cellcolor{deltaloss!29}$-$0.08 & \cellcolor{deltaloss!27}$-$0.07 \\
Reorder arguments & \cellcolor{deltaloss!25}$-$0.06 & \cellcolor{deltaloss!24}$-$0.05 & $-$0.01 & $-$0.01 & \cellcolor{deltaloss!23}$-$0.05 & 0.00 & \cellcolor{deltagain!22}+0.04 & $-$0.01 & 0.00 & \cellcolor{deltaloss!23}$-$0.05 \\
\midrule
\multicolumn{11}{@{}l}{\emph{Cross-call dependency}} \\
Transaction & \cellcolor{deltaloss!32}$-$0.10 & \cellcolor{deltaloss!31}$-$0.10 & $-$0.03 & \cellcolor{deltaloss!22}$-$0.04 & +0.01 & $-$0.02 & \cellcolor{deltaloss!22}$-$0.04 & \cellcolor{deltaloss!25}$-$0.06 & \cellcolor{deltaloss!36}$-$0.12 & \cellcolor{deltaloss!27}$-$0.07 \\
Schema discovery & \cellcolor{deltaloss!28}$-$0.08 & $-$0.01 & +0.02 & \cellcolor{deltaloss!24}$-$0.05 & \cellcolor{deltaloss!30}$-$0.09 & \cellcolor{deltagain!28}+0.08 & \cellcolor{deltaloss!40}$-$0.15 & \cellcolor{deltaloss!67}$-$0.31 & \cellcolor{deltaloss!36}$-$0.12 & \cellcolor{deltaloss!38}$-$0.14 \\
\midrule
\multicolumn{11}{@{}l}{\emph{Composition}} \\
Composed & +0.02 & \cellcolor{deltagain!21}+0.04 & \cellcolor{deltaloss!68}$-$0.32 & \cellcolor{deltaloss!60}$-$0.27 & \cellcolor{deltaloss!59}$-$0.26 & \cellcolor{deltagain!38}+0.14 & \cellcolor{deltaloss!44}$-$0.18 & \cellcolor{deltaloss!66}$-$0.30 & \cellcolor{deltaloss!37}$-$0.13 & \cellcolor{deltaloss!36}$-$0.12 \\
\bottomrule
\end{tabular}}
\end{center}
\end{table}

Table~\ref{tab:real_proxy} tests schema bias on four real benchmarks with
eleven of the twelve representative variants. Reference resolution is not
run, because its value tables come from task data. The runs use the schema
proxy and the task sets of Appendix~\ref{sec:app-benchmarks}, with Qwen3.5-27B
and Qwen3-4B as the models under test.

The costly variants are the same on every benchmark except \tautwo{} airline.
For Qwen3.5-27B, fully merged is the worst variant on retail, BFCL, and
AutomationBench ($-0.30$, $-0.67$, and $-0.44$ from native), and class dispatch
on MCP-Atlas ($-0.40$). Composed costs $0.27$ to $0.32$ on BFCL,
AutomationBench, and MCP-Atlas, class dispatch $0.42$ on BFCL and $0.29$ on
AutomationBench, and interval split $0.23$ on AutomationBench. Qwen3-4B loses
more, in proportion to a lower native score. Fully merged, class dispatch,
schema discovery, and composed take it from $0.33$ to between $0.00$ and $0.03$
on BFCL. On AutomationBench, fully merged takes it from $0.17$ to $0.00$, and
schema discovery, composed, and transaction to $0.03$ or $0.04$. On MCP-Atlas,
these four variants cost it $0.12$ to $0.14$ of $0.17$. The variants that change
only a tool's own representation stay close to native for both models: nested
args, reorder arguments, and fully split are within $0.10$ of native on every
benchmark. Strip descriptions is the exception on BFCL ($-0.23$ for
Qwen3.5-27B and $-0.10$ for Qwen3-4B), whose tools rely on their descriptions.
\tautwo{} airline separates the variants less than the other benchmarks. Apart
from fully merged for Qwen3.5-27B ($-0.16$), the changes lie within $0.14$ of
native, against a standard error of $0.06$ to $0.07$ on 50 tasks.

\FloatBarrier

\section{Additional results on failure modes (RQ2)}
\label{sec:app-rq2}

\subsection{Schema-form gaps}
\label{sec:app-schema-form}

\begin{figure}[t]
\centering
\includegraphics[width=0.84\linewidth]{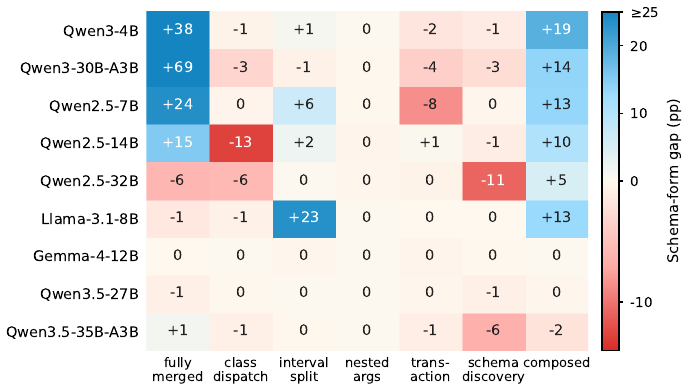}  
\caption{\textbf{Schema-form gap per model and variant.} Change in success
(percentage points) when off-schema calls are executed as the native action
they decode to rather than rejected. Positive values (blue) are schema-form
failures; negative values (red) mean rejection with retry does better than
executing the decoded call. The colour scale saturates at $+25$; the printed
number is the true gap. Nested args is a near-zero control; interval split
uses one threshold.}
\label{fig:rq2_schema_form}
\end{figure}

Figure~\ref{fig:rq2_schema_form} compares two scorings of the same episodes:
invalid calls rejected (the default) or executed as the native call they decode
to. The gap is largest on fully merged: 69 points for Qwen3-30B-A3B, 38 for
Qwen3-4B, 24 for Qwen2.5-7B, and 15 for Qwen2.5-14B. On Qwen3-4B, 544 of the
2{,}748 composed-variant episodes succeed when invalid calls are executed, and
every one of them decodes to the gold native names; under fully merged, 66\% of
calls are still rejected, typically because the model calls a native name
instead of the merged tool. On composed variants the gap is 5 to 19 points for
the earlier-generation models except Gemma-4-12B, whose gap is near zero.
Llama-3.1-8B gains 23 points on interval split, where the other models stay
within 7 points. Negative gaps, where rejection followed by a retry does better,
are largest for Qwen2.5-14B on class dispatch ($-13$), Qwen2.5-32B on schema
discovery ($-11$), and Qwen3.5-35B-A3B on schema discovery ($-6$). No
newer-generation or Gemma cell has a positive gap above 2 points.

\subsection{Audit of the failure labels}
\label{sec:app-audit}

The seven failure modes of Table~\ref{tab:signatures} are assigned by rules,
without manual annotation. To bound their reliability, we sample 150 failed
episodes approximately evenly across model--variant strata, remove the model,
variant, outcome, automatic label, and decoder flags, and ask three independent
LLM labelers (Qwen3, Qwen2.5, and Llama) to assign one of the seven modes to
each episode. This is an audit against LLM proxies, not human validation.

A majority label (at least two of the three labelers agree) exists for 88\% of
the episodes. Two modes cannot be recovered by the labelers: transaction handle
and rejected-and-stuck are defined by decoder flags that the blinded episodes
omit, and no labeler ever chose them, although 45 of the 150 automatic labels
fall in these two modes. On the majority-labelled episodes whose automatic label
is one of the other five modes, the automatic label matches the majority in
77\%. We therefore use the modes to compare aggregate error distributions rather
than to interpret individual episodes.
\FloatBarrier

\section{Estimating schema difficulty (RQ3)}
\label{sec:app-rq3}

Every estimator scores each variant for a model, and $\rho$ is the Spearman
correlation between these scores and the full 2{,}748-query success rates,
computed within each model over the variants that estimator scores and then
averaged over the nine models; scores are never pooled across models.
Table~\ref{tab:rq3_estimators} summarizes every estimator: what it needs, what it costs per
variant, and how well it ranks the variants.

\begin{table}[t]
\caption{\textbf{Performance and cost of different estimators.} Each estimator scores every variant for one model,
and $\rho$ is the Spearman correlation between these scores and the model's full score over the variants, averaged
over models (synthetic) or model--benchmark pairs (real). \emph{Synthetic}: 2{,}748 task queries, nine models. \emph{Real}: the four real benchmarks of
Appendix~\ref{sec:app-real-proxy}, over the model--benchmark pairs whose variant ranking is stable (two random halves
of the tasks agree with $\rho\geq0.65$): 7 pairs for Qwen3.5-27B and Qwen3-4B, and 2 for Qwen3-4B
after RL on the mixture (Table~\ref{tab:rq4_real}). \emph{Exact call} counts a probe as passed only if the call
equals the instructed one. Samples are drawn 40 times per size; the real benchmarks are too small for 100 and 800.
\emph{Split half} scores half of the tasks against the other half, an upper reference for any sample.}
\label{tab:rq3_estimators}
\begin{center}
\scriptsize
\begin{tabular*}{\linewidth}{@{\extracolsep{\fill}}l c l c c c@{}}
\toprule
& & & \multicolumn{3}{c}{Spearman $\rho$ with the full score} \\
\cmidrule(l){4-6}
Estimator & Tasks used & Cost per variant & Synthetic & Real & Real, after RL \\
\midrule
Schema compliance probe & no & 36 instructions & 0.57 & 0.58 & 0.57 \\
\quad exact call & no & 36 instructions & -- & 0.72 & 0.48 \\
Likelihood of the correct calls & no & 36 scored sequences & 0.52 & 0.55 & $-$0.35 \\
\midrule
Sample of 36 task queries & yes & 36 tasks & 0.72 & 0.82 & 0.84 \\
Sample of 100 task queries & yes & 100 tasks & 0.81 & -- & -- \\
Sample of 800 task queries & yes & 800 tasks & 0.92 & -- & -- \\
Split half (reference) & yes & half of the tasks & -- & 0.84 & 0.93 \\
Full evaluation & yes & all tasks & 1.00 & 1.00 & 1.00 \\
\bottomrule
\end{tabular*}
\end{center}
\end{table}

\textbf{Probes that do not use the task queries.} The schema compliance probe
runs 36 single-operation instructions per variant, derived from the schema
(three per domain, no task queries), and records compliance, the fraction of
probes that execute at least one native action with no rejection. On the 13
probed variants, compliance ranks each model's variants with mean $\rho=0.57$
(range $0.00$ to $0.88$). The likelihood probe converts the same instructions'
gold native calls into each variant's correct call sequence, executes it to
obtain the tool responses, and scores the log-probability the model assigns to
its own turns without generating; ranked by this score, $\rho=0.52$ (range
$0.24$ to $0.75$). The probes fall short because following a schema on isolated calls does not
predict following it inside tasks:
six of the seven earlier-generation models follow the schema-discovery protocol
on isolated calls (compliance $0.67$--$1.00$) yet score $0.03$--$0.29$ on the
task queries, and Qwen3-4B follows the transaction protocol in isolation
($0.97$) but scores $0.28$ on tasks; conversely, the Qwen2.5 models fail the
isolated transaction probe ($0.00$--$0.03$), while Qwen2.5-32B scores $0.77$ on
transaction tasks. The newer-generation models comply on almost every probe and
also score $0.75$ to $0.84$ on the corresponding tasks.

\textbf{Samples of the target queries.} Subsampling a variant's queries (40
draws per size) ranks a model's variants with $\rho=0.72$ at 36 queries, $0.81$
at 100, and $0.92$ at 800; the lower end of each range is set by the
newer-generation models, whose variants differ by only a few points. The
correlation uses every variant with episode-level logs (27 for Qwen3-4B and the
newer-generation models, 17--19 for the others).

\textbf{On the four real benchmarks.} The last two columns of
Table~\ref{tab:rq3_estimators} repeat the
comparison on the runs of Appendix~\ref{sec:app-real-proxy}, with the model
trained by RL on the mixture as a third model (Appendix~\ref{sec:app-transfer}).
Each estimator ranks one model's variants on one benchmark. The probes use 36
instructions derived from each benchmark's own tool catalog, built as on the
synthetic benchmark. The ranking can only be estimated where it is stable, so
we report the cells where two random halves of the tasks agree with
$\rho\geq0.65$.

In those nine cells, a sample of 36 tasks reaches $\rho=0.73$ to $0.92$, within
$0.05$ of the split-half agreement in seven of them. For the two untrained models,
the most useful query-free probe is the compliance probe restricted to calls
that equal the instructed one ($\rho=0.64$ to $0.87$). Plain compliance saturates
for Qwen3.5-27B ($\rho=0.36$ to $0.51$). After RL on the mixture, both probes
fail. The trained model passes almost every isolated probe (mean compliance
$0.97$ to $0.99$ per benchmark), and the likelihood of its correct calls is
anti-correlated with task difficulty ($\rho=-0.54$ on BFCL and $-0.15$ on
AutomationBench). Training raised isolated-call behaviour on the trained variants
without a matching gain on tasks (Table~\ref{tab:rq4_real}). A probe calibrated
on a model should therefore not be reused after that model is trained.

\FloatBarrier

\section{Reducing schema bias through training (RQ4)}
\label{sec:app-rq4}

\subsection{Training-free methods}
\label{sec:app-training-free}

Table~\ref{tab:rq4-icl} lists the one-sentence instruction used for each
variant and the decoding setting. On models other than Qwen3-4B, the
schema-discovery instruction adds about 45 points for the models that follow it
and almost nothing for the others. Constraining the first call raises
Gemma-4-12B on fully merged from $0.00$ to $0.50$ and Qwen3-30B-A3B from $0.07$
to $0.74$, but lowers Qwen3.5-4B by 12 points. The transaction instruction adds
at most 10 points (Qwen2.5-7B from $0.25$ to $0.35$, Qwen2.5-14B $+8$) and
changes the remaining models by at most 4 points.

\begin{table}[t]
\caption{Training-free interventions used in \S\ref{sec:rq4}. \emph{Instruction}
appends the listed sentence for each variant to the default system prompt,
which already requires using only the provided tools; the native schema needs
no instruction. \emph{Decoding} sets vLLM \texttt{tool\_choice=required} on the
first turn only, so the first completion is grammar-constrained to a
schema-valid tool name and arguments; later turns use the default
\texttt{auto} policy.}
\label{tab:rq4-icl}
\begin{center}
\scriptsize
\begin{tabularx}{\linewidth}{@{}l X@{}}
\toprule
Variant & Instruction \\
\midrule
Fully merged & All operations go through a single dispatcher tool. Set \texttt{operation} to the operation you need and pass each of its arguments under the key \texttt{\textless{}operation\textgreater{}::\textless{}argument\textgreater{}}. \\
Class dispatch & Each class has one dispatcher tool named after the class. Call the tool of the right class, set \texttt{operation} to the method you need, and pass each argument under the key \texttt{\textless{}method\textgreater{}::\textless{}argument\textgreater{}}. \\
Fully split & The values of enumerable arguments are part of the tool names. Call the tool whose name contains the values you need and pass only the remaining arguments. \\
Interval split & Numeric arguments are split into ranges across several tools. Call the tool whose name gives the range that contains your value, and still pass the value itself. \\
Nested args & Some arguments are grouped into nested objects. Put each argument inside the object that the tool's parameter schema defines for it. \\
Namespaced names & Tool names are prefixed with their class name. Call each tool by its full prefixed name, using the class you intend to act on. \\
Strip descriptions & The tools come without descriptions. Infer what each tool does from its name and its argument names. \\
Reorder arguments & The order of arguments in the tool definitions is arbitrary. Fill every argument by its name, not by its position. \\
Transaction & Open a transaction and read the handle it returns, pass that exact handle to every subsequent write, and finish by confirming the transaction. Do not invent or reuse handles, and do not leave a transaction unconfirmed. \\
Reference resolution & Some arguments take a handle instead of a raw value. First call the matching resolver tool with the user's value, then pass the handle it returns. \\
Schema discovery & Retrieve a class's available methods and argument names before the first call to that class, and use only the returned names. Do not guess argument names. \\
\bottomrule
\end{tabularx}
\end{center}
\end{table}

\subsection{Training setup}
\label{sec:app-train}

All training uses Qwen3-4B-Instruct-2507 and the same 2{,}000 queries from the
eight training domains (Table~\ref{tab:synthetic_domains}), phrased as
templated requests. They are sampled afresh with a different seed, and any
query whose domain and gold-call multiset match an evaluation query is
rejected, so no evaluation task appears in training. SFT traces are oracle call sequences for these queries
under the training variant, replayed through the environment to check that they
reproduce the gold native actions with no rejection. RL uses the same queries
as prompts; the reward is exact success plus partial credit for recalled native
actions, minus penalties for spurious actions and protocol violations.
SFT uses LoRA \citep{hu2022lora} (rank 32, $\alpha=64$) on all attention and MLP
projections, with the loss on assistant turns only, two epochs, learning rate
$2{\times}10^{-4}$, and a maximum sequence length of 16k tokens (32k for the
seven-variant mixture). RL uses on-policy GRPO \citep{shao2024deepseekmath} in
slime \citep{thudm2025slime}, with learning rate $1{\times}10^{-6}$, eight
samples per prompt, a global batch of 96 episodes, KL coefficient $0$, $0.01$, or
$0.1$, rollout temperature $1.0$, and 100 rollouts per run.

\subsection{What training data, KL penalty, and feedback contribute}
\label{sec:app-ablation}

Table~\ref{tab:rq4-ablation} collects the ablation runs behind \S\ref{sec:rq4}.
These ablations were run on an earlier six-variant mixture (fully merged,
transaction, reference resolution, schema discovery, opaque names, and a
one-argument enum split); the \emph{Mixed} columns of
Table~\ref{tab:rq4-methods} use the seven-variant mixture described in
\S\ref{sec:rq4}.

\textbf{SFT.} Training on one variant repairs it: transaction-only data lift
transaction from $0.28$ to $0.57$, and schema-discovery-only data lift schema
discovery from $0.07$ to $0.74$. The tax on other variants depends on the data:
transaction-only data leave native unchanged and also lift fully merged to
$0.75$, whereas schema-discovery-only data lower native by 16 points and
reference resolution by 21. Native-only data leave the hard variants unrepaired.
Seeds differ widely on mixed data (native $0.60$, $0.70$, and $0.81$ for the
three seeds of the six-variant mixture, and $0.42$ to $0.84$ for the four seeds
of the seven-variant mixture). On Qwen2.5-7B, which starts farther from its ceiling, the same
mixed data raise every evaluated variant (native from $0.54$ to $0.78$, two
seeds).

\textbf{RL.} On transaction data, GRPO raises the reward from $-0.22$ to near
its ceiling within about fifteen updates and holds it for the remaining
updates, with no truncated episodes. Native-only and mixed RL keep every
already-robust variant at or above its untrained score; transaction-only RL
with KL$=0$ is the exception, lowering reference resolution from $0.77$ to
$0.63$. Repairing schema discovery needs
both the variant in the data and no KL penalty: mixed data with KL$=0$ reach
$0.75$, while mixed data with KL$=0.01$ reach $0.25$, schema-discovery-only data
with KL$=0.01$ reach $0.12$, and transaction-only data with KL$=0$ reach
$0.20$. The rejection signal, not its wording, drives learning: on
transaction-only data, the schema-discovery variant, which is not trained,
reaches $0.49$ with explained rejections and $0.49$ with bare refusals, but only
$0.09$ when invalid calls are executed instead of rejected. Starting GRPO from
the mixed SFT adapter reaches the same scores as mixed RL from the untrained
model even with KL$=0.01$, because SFT has already moved the model away from
its initialization.

\begin{table}[t]
\caption{Training ablations on Qwen3-4B: exact success on the full 2{,}748-query
set for six of the twelve representative variants, by training method, training
data, and KL coefficient ($n$ = seeds averaged). \emph{Mixed} data rotate over
fully merged, transaction, reference resolution, schema discovery, opaque names,
and a one-argument enum split. The last three RL rows change the environment or
the reward: rejections without an explanation, no rejections (invalid calls are
executed as their decoded native call), and a reward of exact success only.
The SFT$\to$RL rows start GRPO from the mixed SFT adapter.}
\label{tab:rq4-ablation}
\begin{center}
\footnotesize
\setlength{\tabcolsep}{4pt}
\resizebox{\linewidth}{!}{%
\begin{tabular}{@{}lccccccc@{}}
\toprule
Training data & $n$ & Native & Fully split & Fully merged & Ref.\ res. & Trans. & Schema disc. \\
\midrule
Untrained &  & 0.77 & 0.76 & 0.37 & 0.77 & 0.28 & 0.07 \\
\midrule
\multicolumn{1}{@{}l}{\emph{SFT}} & &  &  &  &  &  &  \\
~~native only & 2 & 0.79 & 0.81 & 0.47 & 0.64 & 0.18 & 0.02 \\
~~transaction only & 2 & 0.81 & 0.79 & 0.75 & 0.83 & 0.57 & 0.35 \\
~~schema discovery only & 2 & 0.61 & 0.55 & 0.59 & 0.57 & 0.39 & 0.74 \\
~~mixed & 3 & 0.70 & 0.71 & 0.69 & 0.74 & 0.50 & 0.70 \\
\midrule
\multicolumn{1}{@{}l}{\emph{RL, KL$=0$}} & &  &  &  &  &  &  \\
~~native only & 1 & 0.83 & 0.82 & 0.48 & 0.83 & 0.30 & 0.13 \\
~~transaction only & 2 & 0.78 & 0.76 & 0.55 & 0.63 & 0.77 & 0.20 \\
~~mixed & 2 & 0.81 & 0.80 & 0.79 & 0.87 & 0.65 & 0.75 \\
\midrule
\multicolumn{1}{@{}l}{\emph{RL, KL$=0.01$}} & &  &  &  &  &  &  \\
~~transaction only & 1 & 0.77 & 0.76 & 0.57 & 0.81 & 0.74 & 0.49 \\
~~schema discovery only & 1 & 0.81 & 0.81 & 0.68 & 0.74 & 0.37 & 0.12 \\
~~mixed & 2 & 0.82 & 0.82 & 0.80 & 0.88 & 0.54 & 0.25 \\
~~transaction only, bare refusals & 1 & 0.77 & 0.74 & 0.48 & 0.81 & 0.75 & 0.49 \\
~~transaction only, no rejections & 1 & 0.79 & 0.78 & 0.46 & 0.81 & 0.76 & 0.08 \\
~~transaction only, exact-match reward & 1 & 0.78 & 0.77 & 0.50 & 0.66 & 0.72 & 0.11 \\
\midrule
\multicolumn{1}{@{}l}{\emph{SFT$\to$RL}} & &  &  &  &  &  &  \\
~~mixed, KL$=0.01$ & 2 & 0.81 & 0.80 & 0.79 & 0.86 & 0.67 & 0.75 \\
\bottomrule
\end{tabular}}
\end{center}
\end{table}

\subsection{Phrasing and volume of SFT data}
\label{sec:app-phrasing}

To separate how demonstrations are phrased from how many there are, we train
LoRA adapters on templated or colloquially rephrased versions of the same
traces (same gold calls) at 500, 1{,}000, 2{,}000, and 4{,}000 examples, with
two seeds each, and evaluate them on ten variants in the four held-out domains
(160 evaluations). Colloquial data win at every size: the mean over the ten
variants is $0.680$ versus $0.447$ at 500 traces, $0.711$ versus $0.569$ at
1{,}000, $0.698$ versus $0.473$ at 2{,}000, and $0.691$ versus $0.607$ at
4{,}000. More data do not help beyond 1{,}000 colloquial traces.

\subsection{Transfer}
\label{sec:app-transfer}

Table~\ref{tab:transfer-summary} summarizes three transfer tests.

\textbf{Held-out domains.} With the seven-variant mixture, SFT keeps on average
43\% of its training-domain gain on transaction (8 to 61\% across four seeds) and
67\% on schema discovery in the four held-out domains, while RL keeps 65\% and
71\% (three seeds). Held-out transaction success reaches $0.38$ after SFT and
$0.49$ after RL.

\textbf{Held-out schema changes.} A separate LoRA adapter is trained on 2{,}000
traces rotating over seven changes (fully merged, one-cut interval split, nested
args, transaction, reference resolution, opaque names, and a one-argument enum
split) and evaluated on changes that were not trained: new combinations of the
trained changes (transaction composed with reference resolution, and a catalog
mixing several operators), a stronger version of a trained change (interval
split with four cut points instead of one), new operators (splitting every enum
argument, and semantic grouped merges), and new class-level structures (class
dispatch and schema discovery). Relative to the untrained model, the trained
changes improve by 14.6 points, new combinations by 8.8, the stronger version by
11.2, new operators by $-2.4$, and new class-level structures by 1.2 (exact
success; recall $-3.5$).

\textbf{Real benchmarks.} We evaluate the RL Mixed model of Table~\ref{tab:rq4-methods} (seed 42) on
the four benchmarks of Appendix~\ref{sec:app-real-proxy}. Table~\ref{tab:rq4_real}
compares it with the untrained model, variant by variant. The gains stay on the
trained variants. Fully merged improves on retail ($+0.21$) and AutomationBench
($+0.12$), and schema discovery by $0.02$ to $0.15$ on four of the five
benchmarks. Across all trained variants other than native, the mean change is
$+0.03$. The variants outside the training mixture change by $+0.003$ on average,
and 24 of their 29 cells move by at most $0.05$. Native itself, which is in the
mixture, rises on \tautwo{} airline and retail ($+0.12$ and $+0.08$), where the
trained model is also the user simulator, and drops on BFCL, AutomationBench, and
MCP-Atlas ($-0.01$, $-0.03$, and $-0.10$). Training on a mixture of schema
variants thus repairs specific trained conventions in real environments, but
does not make the model robust to schema changes in general.

\begin{table}[t]
\caption{\textbf{Training mitigation barely transfers to real benchmarks.} RL on mixed-variant data, evaluated on
the real benchmarks: Qwen3-4B after GRPO on the seven-variant mixture
(the RL Mixed column of Table~\ref{tab:rq4-methods}, seed 42) minus the untrained model, per variant and benchmark;
shading as in Table~\ref{tab:rq4-methods}. Red frames mark the variants in the training mixture (reference resolution,
also in it, is not run on these benchmarks). Benchmarks and scores as in Table~\ref{tab:real_proxy}.}
\label{tab:rq4_real}
\begin{center}
\scriptsize
\begin{tabular*}{\linewidth}{@{\extracolsep{\fill}}l ccccc@{}}
\toprule
Variant & $\tau^2$ air. & $\tau^2$ ret. & BFCL & AutoB. & MCP-A. \\
\midrule
Native & \cellcolor{deltagain!35}\setlength{\fboxsep}{1pt}\setlength{\fboxrule}{0.7pt}\fcolorbox{trainframe}{deltagain!35}{+0.12} & \cellcolor{deltagain!28}\setlength{\fboxsep}{1pt}\setlength{\fboxrule}{0.7pt}\fcolorbox{trainframe}{deltagain!28}{+0.08} & \setlength{\fboxsep}{1pt}\setlength{\fboxrule}{0.7pt}\fcolorbox{trainframe}{white}{$-$0.01} & \cellcolor{deltaloss!20}\setlength{\fboxsep}{1pt}\setlength{\fboxrule}{0.7pt}\fcolorbox{trainframe}{deltaloss!20}{$-$0.03} & \cellcolor{deltaloss!32}\setlength{\fboxsep}{1pt}\setlength{\fboxrule}{0.7pt}\fcolorbox{trainframe}{deltaloss!32}{$-$0.10} \\
\midrule
\multicolumn{6}{@{}l}{\emph{Tool-set partitioning}} \\
Fully merged & \cellcolor{deltaloss!25}\setlength{\fboxsep}{1pt}\setlength{\fboxrule}{0.7pt}\fcolorbox{trainframe}{deltaloss!25}{$-$0.06} & \cellcolor{deltagain!50}\setlength{\fboxsep}{1pt}\setlength{\fboxrule}{0.7pt}\fcolorbox{trainframe}{deltagain!50}{+0.21} & \setlength{\fboxsep}{1pt}\setlength{\fboxrule}{0.7pt}\fcolorbox{trainframe}{white}{+0.01} & \cellcolor{deltagain!34}\setlength{\fboxsep}{1pt}\setlength{\fboxrule}{0.7pt}\fcolorbox{trainframe}{deltagain!34}{+0.12} & \setlength{\fboxsep}{1pt}\setlength{\fboxrule}{0.7pt}\fcolorbox{trainframe}{white}{+0.01} \\
Class dispatch & \cellcolor{deltagain!32}+0.10 & +0.03 & +0.01 & +0.03 & $-$0.03 \\
Fully split & \cellcolor{deltagain!28}\setlength{\fboxsep}{1pt}\setlength{\fboxrule}{0.7pt}\fcolorbox{trainframe}{deltagain!28}{+0.08} & \cellcolor{deltagain!24}\setlength{\fboxsep}{1pt}\setlength{\fboxrule}{0.7pt}\fcolorbox{trainframe}{deltagain!24}{+0.05} & \setlength{\fboxsep}{1pt}\setlength{\fboxrule}{0.7pt}\fcolorbox{trainframe}{white}{$-$0.01} & \setlength{\fboxsep}{1pt}\setlength{\fboxrule}{0.7pt}\fcolorbox{trainframe}{white}{0.00} & \cellcolor{deltaloss!21}\setlength{\fboxsep}{1pt}\setlength{\fboxrule}{0.7pt}\fcolorbox{trainframe}{deltaloss!21}{$-$0.03} \\
Interval split & $-$0.02 & n/a & +0.02 & \cellcolor{deltaloss!23}$-$0.05 & \cellcolor{deltaloss!22}$-$0.04 \\
\midrule
\multicolumn{6}{@{}l}{\emph{Per-tool representation}} \\
Nested args & \cellcolor{deltaloss!25}$-$0.06 & 0.00 & \cellcolor{deltagain!22}+0.04 & +0.01 & \cellcolor{deltaloss!29}$-$0.08 \\
Namespaced names & \setlength{\fboxsep}{1pt}\setlength{\fboxrule}{0.7pt}\fcolorbox{trainframe}{white}{0.00} & \cellcolor{deltagain!30}\setlength{\fboxsep}{1pt}\setlength{\fboxrule}{0.7pt}\fcolorbox{trainframe}{deltagain!30}{+0.09} & \setlength{\fboxsep}{1pt}\setlength{\fboxrule}{0.7pt}\fcolorbox{trainframe}{white}{+0.02} & \setlength{\fboxsep}{1pt}\setlength{\fboxrule}{0.7pt}\fcolorbox{trainframe}{white}{$-$0.02} & \cellcolor{deltaloss!24}\setlength{\fboxsep}{1pt}\setlength{\fboxrule}{0.7pt}\fcolorbox{trainframe}{deltaloss!24}{$-$0.06} \\
Strip descriptions & \cellcolor{deltagain!25}+0.06 & +0.01 & $-$0.02 & +0.01 & \cellcolor{deltaloss!20}$-$0.03 \\
Reorder arguments & \cellcolor{deltagain!42}+0.16 & +0.02 & \cellcolor{deltaloss!21}$-$0.04 & $-$0.02 & \cellcolor{deltaloss!22}$-$0.04 \\
\midrule
\multicolumn{6}{@{}l}{\emph{Cross-call dependency}} \\
Transaction & \cellcolor{deltagain!25}\setlength{\fboxsep}{1pt}\setlength{\fboxrule}{0.7pt}\fcolorbox{trainframe}{deltagain!25}{+0.06} & \setlength{\fboxsep}{1pt}\setlength{\fboxrule}{0.7pt}\fcolorbox{trainframe}{white}{0.00} & \setlength{\fboxsep}{1pt}\setlength{\fboxrule}{0.7pt}\fcolorbox{trainframe}{white}{$-$0.02} & \cellcolor{deltagain!21}\setlength{\fboxsep}{1pt}\setlength{\fboxrule}{0.7pt}\fcolorbox{trainframe}{deltagain!21}{+0.04} & \setlength{\fboxsep}{1pt}\setlength{\fboxrule}{0.7pt}\fcolorbox{trainframe}{white}{$-$0.02} \\
Schema discovery & \setlength{\fboxsep}{1pt}\setlength{\fboxrule}{0.7pt}\fcolorbox{trainframe}{white}{0.00} & \cellcolor{deltagain!40}\setlength{\fboxsep}{1pt}\setlength{\fboxrule}{0.7pt}\fcolorbox{trainframe}{deltagain!40}{+0.15} & \cellcolor{deltagain!28}\setlength{\fboxsep}{1pt}\setlength{\fboxrule}{0.7pt}\fcolorbox{trainframe}{deltagain!28}{+0.07} & \cellcolor{deltagain!20}\setlength{\fboxsep}{1pt}\setlength{\fboxrule}{0.7pt}\fcolorbox{trainframe}{deltagain!20}{+0.03} & \setlength{\fboxsep}{1pt}\setlength{\fboxrule}{0.7pt}\fcolorbox{trainframe}{white}{+0.02} \\
\midrule
\multicolumn{6}{@{}l}{\emph{Composition}} \\
Composed & 0.00 & 0.00 & 0.00 & 0.00 & +0.01 \\
\bottomrule
\end{tabular*}
\end{center}
\end{table}

\begin{table}[t]
\caption{Transfer of training gains on Qwen3-4B
(Appendix~\ref{sec:app-transfer}). Changes are in exact success relative to the
untrained model; the last row uses each benchmark's own score.}
\label{tab:transfer-summary}
\begin{center}
\small
\begin{tabular}{@{}p{0.17\linewidth}p{0.33\linewidth}p{0.44\linewidth}@{}}
\toprule
held out & evaluated on & result \\
\midrule
Domains &
  transaction and schema discovery, four held-out domains &
  SFT keeps 43\% (transaction) and 67\% (schema discovery) of its gain; RL keeps 65\% and 71\% \\
Schema changes (SFT) &
  new combination / stronger version / new operator / class-level structure &
  $+8.8$ / $+11.2$ / $-2.4$ / $+1.2$ points (trained changes $+14.6$) \\
Benchmarks &
  four real benchmarks, RL on the seven-variant mixture (Table~\ref{tab:rq4_real}) &
  trained variants $+3.0$, untrained variants $+0.3$ points (mean over variant--benchmark cells) \\
\bottomrule
\end{tabular}
\end{center}
\end{table}

\subsection{Additional training studies}
\label{sec:app-extra-training}

Two further studies test whether training recipes found on the synthetic tasks
hold on other targets. Neither changes the conclusions of \S\ref{sec:rq4}; both
show that the effect of a recipe depends on the target.

\textbf{Interface factors derived from failures.} We isolate three interface
factors behind recurring failures on \tautwo{}, each holding the task semantics
fixed: whether a policy constraint is stated in the system prompt or next to the
tool it governs; whether an entity is found with one composite lookup or a chain
of three lookups; and whether the tool list shows every state transition or only
those currently legal. On Qwen3-4B (160 tasks per factor, five realizations
each), placing the policy next to the tool has no effect (0.0 points), chained
lookups lower exact success by 86.1 points ($[-88.5, -83.6]$), and showing only
legal transitions raises it by 8.8 points ($[6.5, 11.0]$). Training on the two
effective factors (GRPO, three $2{\times}2$ designs, three seeds, 36 runs) shows
that factors do not combine additively: covering the relevant task semantics and
exposing the new interface interact negatively ($-17.6$ points,
$[-27.0, -9.1]$), and none of the corresponding effects on \tautwo{} excludes
zero.

\textbf{Training for an airline domain.} We train Qwen3-4B on a generated airline
environment (48 runs across three designs: training-data properties, interface
choices, and training order) and evaluate on held-out generated airline tasks
and on \tautwo{} airline. Multi-turn training data lower held-out success by
14.5 points ($[-22.6, -5.2]$) and deeper queries raise it by 4.2 points
($[0.5, 7.9]$); no interface choice has an effect that survives Holm correction
on either target. The best training order reverses between targets: SFT alone
beats RL from the untrained model by 39.6 points ($[22.5, 52.7]$) on the
generated tasks but trails it by 9.7 points ($[-19.5, -3.1]$) on \tautwo{}
airline, where SFT followed by RL beats SFT alone by 10.8 points
($[1.5, 19.0]$). A training recipe should therefore be validated on the target
distribution.
\FloatBarrier

\end{document}